\documentclass[acmsmall]{acmart}
\setcopyright{acmcopyright} 
\setcctype{by}
\acmJournal{PACMCGIT}
\acmYear{2025}
\acmVolume{8}
\acmNumber{1}
\acmArticle{20}
\acmMonth{5}

\acmDOI{10.1145/3728310}

\usepackage{booktabs}
\usepackage{multirow}

\usepackage{enumitem}
\usepackage{bm}
\usepackage{soul}

\usepackage{colortbl}

\usepackage{siunitx}

\usepackage{acmart-taps}

\begin{document}
\title{SplatMAP: Online Dense Monocular SLAM with 3D Gaussian Splatting }
\author{Yue Hu}
\email{yhu57782@usc.edu}
\affiliation{%
  \institution{Institute for Creative Technologies, University of Southern California}
  \city{Los Angeles}
  \state{California}
  \country{USA}
}

\author{Rong Liu}
\email{roliu@ict.usc.edu}
\affiliation{%
  \institution{Institute for Creative Technologies, University of Southern California}
  \city{Los Angeles}
  \state{California}
  \country{USA}
}

\author{Meida Chen}
\email{mechen@ict.usc.edu}
\affiliation{%
  \institution{Institute for Creative Technologies, University of Southern California}
  \city{Los Angeles}
  \state{California}
  \country{USA}
}

\author{Peter Beerel}
\authornote{Co-advisor}
\email{pabeerel@usc.edu}
\affiliation{%
  \institution{University of Southern California}
  \city{Los Angeles}
  \state{California}
  \country{USA}
}

\author{Andrew Feng}
\authornotemark[1]
\email{feng@ict.usc.edu}
\affiliation{%
  \institution{Institute for Creative Technologies, University of Southern California}
  \city{Los Angeles}
  \state{California}
  \country{USA}
}









\begin{abstract}

Achieving high-fidelity 3D reconstruction from monocular video remains challenging due to the inherent limitations of traditional methods like Structure-from-Motion (SfM) and monocular SLAM in accurately capturing scene details. While differentiable rendering techniques such as Neural Radiance Fields (NeRF) address some of these challenges, their high computational costs make them unsuitable for real-time applications. Additionally, existing 3D Gaussian Splatting (3DGS) methods often focus on photometric consistency, neglecting geometric accuracy and failing to exploit SLAM's dynamic depth and pose updates for scene refinement.
We propose a framework integrating dense SLAM with 3DGS for near real-time, high-fidelity dense reconstruction. Our approach introduces \textit{SLAM-Informed Adaptive Densification}, which dynamically updates and densifies the Gaussian model by leveraging dense point clouds from SLAM. Additionally, we incorporate \textit{Geometry-Guided Optimization}, which combines edge-aware geometric constraints and photometric consistency to jointly optimize appearance and geometry of the 3DGS scene representation, enabling detailed and accurate SLAM mapping reconstruction.
Experiments on the Replica and TUM-RGBD datasets demonstrate the effectiveness of our approach, achieving state-of-the-art results among monocular systems. Specifically, our method achieves a PSNR of 36.864, SSIM of 0.985, and LPIPS of 0.040 on Replica, representing improvements of 10.7\%, 6.4\%, and 49.4\%, respectively, over the previous SOTA. On TUM-RGBD, our method outperforms the closest baseline by 10.2\%, 6.6\%, and 34.7\% in the same metrics. These results highlight the potential of our framework in bridging the gap between photometric and geometric dense 3D scene representations, paving the way for practical and efficient monocular dense reconstruction. A demonstration of the results can be found in the accompanying video: \url{https://youtu.be/Pr_kyWQQkGo}.

\end{abstract}



\maketitle
 
\section{Introduction}
\label{sec:intro}
\begin{figure*}[!ht]
    \centering
    \includegraphics[width=.9\linewidth]{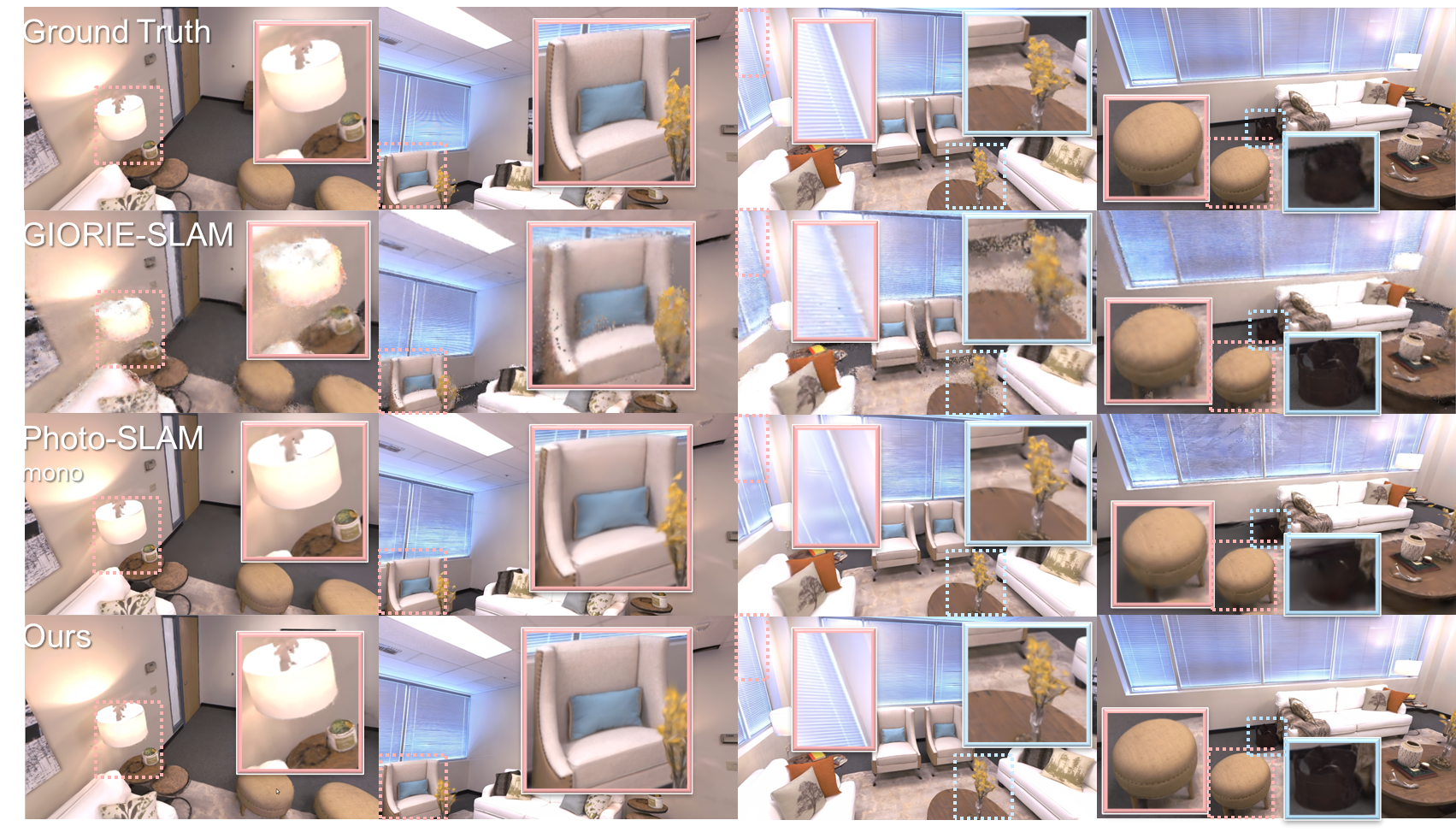}
\caption{Visual comparison of reconstructed scenes. Our method (SplatMap) achieves superior rendering fidelity and geometric accuracy, closely matching the ground truth, while outperforming GIORIE-SLAM~\cite{zhang2024glorie} and Photo-SLAM~\cite{huang2024photo} in fine structural details.}
    \label{fig:vis_comp_abstract}
\end{figure*}
Reconstructing high-fidelity 3D environments from 2D image sequences has been a long-standing goal in computer vision, with applications spanning robotics, augmented reality, and autonomous navigation. Two foundational techniques, Structure-from-Motion (SfM) and Simultaneous Localization and Mapping (SLAM), have played pivotal roles in advancing this field. SfM algorithms, such as COLMAP~\cite{fisher2021colmap} and GLOMAP~\cite{pan2024global}, better at estimating camera parameters from uncalibrated image collections, although at the cost of computational efficiency. In contrast, SLAM systems like DROID-SLAM~\cite{teed2021droid} and ORB-SLAM~\cite{mur2015orb} provide real-time localization and mapping capabilities. Parallel to these developments, differentiable rendering techniques, particularly Neural Radiance Fields (NeRF)~\cite{mildenhall2021nerf}, have revolutionized 3D representation by enabling implicit modeling of appearance and geometry. However, the computational overhead associated with NeRF's rendering processes limits its real-time applicability. Addressing this limitation, 3D Gaussian Splatting (3DGS)~\cite{kerbl20233d} has emerged as a compelling alternative, offering faster rendering and more scalable scene representation.

In order to achieve real-time 3D reconstruction with high-fidelity rendering, researchers have explored the potential of integrating SLAM with differentiable rendering~\cite{zhu2022nice,zhu2024nicer,yang2022vox,keetha2024splatam,johari2023eslamefficientdenseslam,sandstrom2023point,yan2024gsslamdensevisualslam,matsuki2024gaussian,zhang2024glorie,huang2024photo,peng2024rtg}. However, monocular SLAM systems, particularly during early mapping stages, are prone to producing inaccurate point clouds due to limited observations, shallow triangulation baselines, and weak pose constraints. These inaccuracies propagate into downstream mapping tasks, leading to artifacts such as "ghosting" on walls and furniture, especially when observed from untrained viewpoints. Traditional 3DGS densification pipelines typically rely on RGB losses to correct these errors, introducing additional computational overhead as they compensate for the initial inaccuracies.

To address these challenges, we propose SLAM-Informed Adaptive Densification (SIAD), a novel strategy that utilizes SLAM's dynamic updates to refine point clouds in real time. By pruning erroneous points, incorporating new observations, and adapting Gaussian representations based on reliable masks, our method ensures accurate and efficient point cloud refinement without requiring costly post-hoc corrections. This approach bridges the gap between monocular dense SLAM and Gaussian-based scene representations, achieving a geometric level of detail and robustness that surpasses traditional pipelines.

In addition, we introduce Geometry-Guided Optimization, which incorporates edge-aware normal loss and photometric consistency loss to jointly optimize the appearance and geometry of the 3DGS representation. This framework enhances rendering fidelity while preserving fine structural details, particularly around object edges and sharp transitions. By integrating these improvements into a unified pipeline, our {\em SplatMap} framework delivers high-fidelity 3D reconstructions from monocular input, suitable for both real-time and high-quality applications.

The contributions of our {\em SplatMap} system are summarized as follows:
\begin{itemize}
    \item \textbf{SLAM-Informed Adaptive Densification:} We upgrade traditional Gaussian splatting by leveraging dense SLAM outputs, enabling dynamic densification of point clouds for richer scene representation. 
    \item \textbf{Geometry-Guided Optimization:} We introduce a novel loss function that integrates geometric and photometric constraints during mapping, improving both visual quality and structural accuracy.
    \item \textbf{Unified Pipeline for Dense Monocular Reconstruction:} By integrating dense monocular SLAM with 3D Gaussian Splatting, our framework achieves high-fidelity reconstruction from monocular input. 
\end{itemize}

\begin{figure*}
    \centering
    \includegraphics[width=\linewidth]{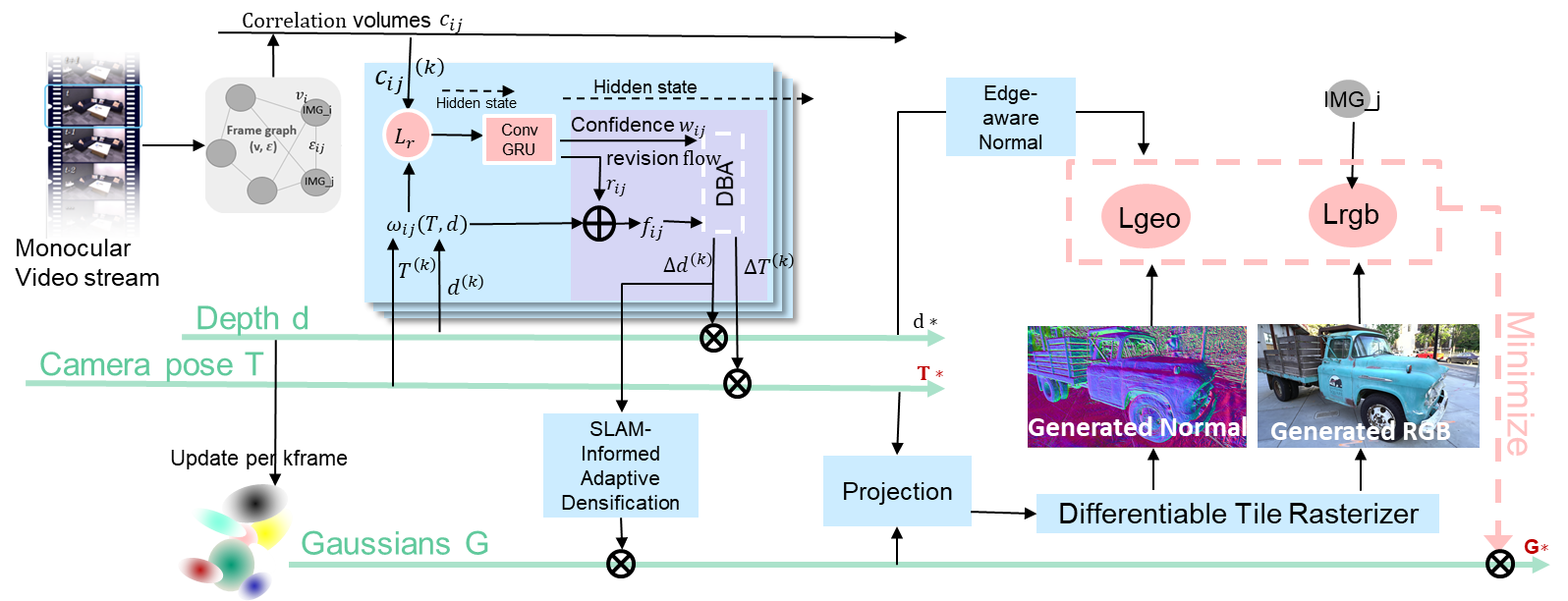}
    \caption{Overview of our proposed system. The framework integrates SLAM-based pose estimation and depth refinement with Gaussian-based 3D scene representations. A convolutional GRU module iteratively refines depth $d$ and pose $T$ using confidence weights $\omega_{ij}$ and revision flow $r_{ij}$. The refined depth and pose are projected to a Gaussian representation $G$, which is optimized to minimize geometric loss ($L_{\text{geo}}$) and photometric loss ($L_{\text{rgb}}$) through a differentiable tile rasterizer. This process generates accurate surface normals and high-quality RGB reconstructions, improving 3DGS SLAM representation fidelity.}
    \label{fig:systemteaser}
\end{figure*}

SplatMap achieves significant improvements over previous monocular SOTA methods, with up to 10.2\% in PSNR, 6.6\% in SSIM, and 34.7\% in LPIPS on real world dataset, demonstrating its effectiveness in refining scene representations across diverse datasets. In Fig.\ref{fig:vis_comp_abstract}, we present a visual comparison between SplatMap and the SOTA rendered mapping results, while Fig.\ref{fig:systemteaser} illustrates the detailed workflow of the system.

\section{Related Work}
In this section, we review prior works relevant to our framework, categorizing them into three major areas: SLAM (dense and sparse), Differentiable Rendering, and SLAM with Differentiable Rendering (RGB-D and monocular). These topics cover foundational techniques, recent advancements, and existing gaps that our proposed SplatMap framework addresses.

\subsection{Simultaneous Localization and Mapping (SLAM)} 
SLAM is a fundamental technique for estimating camera poses and reconstructing 3D environments from image sequences. It can be broadly categorized into sparse SLAM and dense SLAM, depending on how the scene is represented.

\textbf{Sparse SLAM methods} focus on extracting and tracking sparse keypoints across frames, making them computationally efficient and robust for a wide range of scenarios. 
ORB-SLAM~\cite{mur2015orb,mur2017orb,tukan2023orbslam3} has been a cornerstone in SLAM research due to its lightweight and efficient feature-based approach. By extracting and matching features across frames, it achieves robust pose estimation with relatively low computational overhead, making it particularly suitable for both indoor and outdoor environments. Its loop closure mechanisms further enhance long-term mapping consistency by reducing drift over extended trajectories. However, ORB-SLAM's reliance on well-textured scenes limits its effectiveness in scenarios with low texture or repetitive patterns, where feature matching becomes unreliable. As a result, its pose estimation accuracy often falls short in complex or visually ambiguous environments.
LSD-SLAM~\cite{engel2014lsd} represents a significant alternative approach that directly optimizes pixel intensities instead of keypoints, enabling large-scale monocular SLAM. Later, BAD-SLAM~\cite{schops2019bad} combined direct and bundle-adjusted techniques to enhance robustness in RGB-D scenarios. For large-scale 3D reconstruction, volumetric SLAM systems like~\cite{vespa2018efficient} proposed octree-based representations that support both signed-distance and occupancy mapping, offering efficient and scalable solutions.

In contrast, \textbf{dense SLAM methods} aim to reconstruct detailed 3D scene geometry and appearance simultaneously, often leveraging depth maps or pixel-level photometric consistency for optimization.  ElasticFusion~\cite{whelan2015elasticfusion} circumvented traditional pose graph optimization by directly refining dense RGB-D maps, offering an efficient solution for real-time dense SLAM. More recent works like NICE-SLAM~\cite{zhu2022nice} and NICER-SLAM~\cite{zhu2024nicer} introduced neural implicit scene encoding to achieve scalable, memory-efficient dense reconstructions. Real-time large-scale dense RGB-D SLAM~\cite{whelan2015elasticfusion} further improved volumetric fusion, enabling robust reconstruction in complex and dynamic environments. ESLAM~\cite{johari2023eslamefficientdenseslam} leveraged photometric consistency for optimization, enhancing the accuracy of dense reconstructions. Recent innovations like Point-SLAM~\cite{sandstrom2023point} have integrated real-time dense mapping for large-scale scenarios, while Co-SLAM~\cite{wang2023co} jointly optimizes coordinate and sparse parametric encodings to improve reconstruction fidelity. 
Vox-Fusion~\cite{yang2022vox} has pushed the boundaries further by incorporating voxel-based feature maps, providing efficient and detailed scene representations. DROID-SLAM~\cite{teed2021droid} stands out for its use of recurrent neural networks (RNNs) to achieve end-to-end optimization, which makes it particularly robust in estimating both camera pose and depth maps. 
Despite its advancements in camera tracking accuracy and its ability to provide relatively dense point clouds, DROID-SLAM still falls short in enhancing the visual fidelity of the reconstructed maps. Its mapping output often lacks the level of detail and consistency required for high-quality scene representation, leaving room for further improvement in dense 3D reconstruction.

\subsection{Differentiable Rendering}

Recent advances in representing 3D scenes via radiance fields, particularly with Neural Radiance Field (NeRF)~\cite{mildenhall2021nerf}, have enabled implicit modeling of appearance and geometry. Despite improvements in optimization, NeRF methods still suffer from slow rendering due to the neural network querying and volume rendering processes.

Kerbl et al. addressed this by introducing a 3D scene representation using 3D Gaussians and a fast tile-based rasterizer, achieving real-time rendering with high visual quality (3DGS)~\cite{kerbl20233d}. In addition to the aforementioned works, recent efforts by Huang et al. have demonstrated that 2D Gaussian Splatting can markedly improve the geometric accuracy of radiance field reconstruction~\cite{huang20242d}. Their approach refines the rendering process by focusing on geometric fidelity, thereby achieving more accurate and visually consistent outputs. Building on the efficiency of 3D Gaussian Splatting, DreamGaussian~\cite{tang2023dreamgaussian} applied it to generative 3D modeling with mesh extraction and texture refinement for text-to-3D tasks. Wu et al. further extended its use for dynamic scenes, combining 3D Gaussians with 4D neural voxels, enabling efficient feature encoding and Gaussian deformation prediction over time~\cite{wu20244d}.

\subsection{Differentiable Rendering SLAM}

Since the emergence of NeRF, numerous methods have significantly advanced high-fidelity 3D reconstruction using NeRF-based representations. Pioneering works such as iMAP~\cite{sucar2021imap} and Nice-SLAM~\cite{zhu2022nice} introduced NeRF-based dense SLAM systems by fitting radiance fields with Multi-Layer Perceptrons (MLPs) and multi-scale feature grids, respectively. While these approaches enable impressive photorealistic scene reconstructions, they suffer from inherent limitations of NeRF, including high computational costs, slow convergence, and memory inefficiencies, especially for real-time applications. These drawbacks arise from the need to densely sample rays through volumetric representations and the reliance on neural networks for rendering.

Recently, 3DGS has shown great promise in 3D reconstruction by avoiding the aforementioned limitations of NeRF-based representations. It also achieves faster rendering speeds, which is why we chose it as our primary map representation. 
\textbf{Monocular} Photo-SLAM~\cite{huang2024photo} pioneered the integration of photometric consistency into SLAM frameworks, achieving real-time photorealistic mapping from monocular input. GLORIE-SLAM~\cite{zhang2024glorie} advances RGB-only dense SLAM with a neural point cloud representation and a novel DSPO layer for bundle adjustment, jointly optimizing pose, depth, and scale to address the lack of geometric priors in monocular SLAM. This approach improves mapping and rendering accuracy while remaining computationally efficient. For \textbf{RGB-D} input systems, advancements in real-time 3D reconstruction have also been propelled by novel integration strategies of Gaussian splatting techniques. In this context, RTG-slam by Peng et al. has shown promising results by incorporating 3D Gaussian splatting into RGB-D SLAM frameworks~\cite{peng2024rtg}. Their work not only enables efficient large-scale reconstruction but also delivers competitive performance in real-time settings, thereby providing a robust alternative to conventional methods. Other works like SplaTAM~\cite{keetha2024splatam} and GS-SLAM~\cite{yan2024gs} also expand on this by incorporating both depth and color information, using 3DGS for efficient 3D scene representation. These systems enable tasks such as object interaction and robot manipulation, enhancing SLAM applications in robotics.

Building on these insights, our proposed SplatMap framework combines the robust pose estimation capabilities of DROID-SLAM with the high-fidelity mapping strengths of 3D Gaussian Splatting (3DGS). By leveraging SLAM-informed adaptive densification and geometry-guided optimization, SplatMap addresses the limitations of existing methods, achieving both accurate trajectory estimation and visually consistent dense reconstruction.

\section{Methodology}\label{sec:method}
The main idea of our approach is to supervise a 3D Gaussian splatting model using the output from dense monocular SLAM. Dense monocular SLAM can estimate dense depth maps and camera poses, while also providing uncertainty estimates for both depths and poses. With this information, we can train a 3D Gaussian splatting model with a dense depth loss weighted by the depths’ marginal covariances. By using real-time implementations of both dense SLAM and 3D Gaussian splatting training, and by running these in parallel, we achieve real-time performance. Fig.~\ref{fig:systemteaser} shows the flow of information in our pipeline. We now explain our architecture, starting with our tracking frontend and following by our mapping backend. 

It is important to note that our approach builds upon existing DROID-SLAM~\cite{teed2021droid} and 3DGS frameworks. For the sake of completeness, Sec.\ref{sec:tracking} reviews the DROID-SLAM–based tracking module, and Sec.\ref{sec:3dgs1} to ~\ref{sec:3dgs2} describe the 3DGS components inherited from prior work. In contrast, our novel contributions—including the SLAM-Informed Adaptive Densification and Geometry-Guided Optimization are detailed in Sec.\ref{sec:siad} and ~\ref{sec:keyframe}.

\subsection{Video Stream Input and Factor Graph Construction}
\label{sec:Graph}

The SplatMap system takes a continuous monocular video stream as input and incrementally constructs a factor graph that captures the co-visibility relationships between frames. We begin by constructing a factor graph  $G(V, E) $ from the input sequence of video frames, where  $V $ represents the set of frames and  $E $ represents the set of edges between frames. An edge exists between two frames if there are overlapping visual features between them, thereby capturing their co-visibility. As new frames are added, the factor graph is dynamically updated to maintain the co-visibility relationships between frames.

\subsection{Tracking}
\label{sec:tracking}

During tracking, we iteratively refine camera poses  $T $ and depth maps  $d $ by solving an optimization problem defined over the factor graph. Inspired by the approach used in DROID-SLAM~\cite{teed2021droid}, we estimate incremental updates to the camera pose 
($\Delta\xi$) and depth ($\Delta d$) for each frame. 

The optimization process leverages the dense optical flow between frames. Specifically, the predicted flow  $f^{\text{pred}}_{ij} $ represents the pixel displacement field mapping pixels in frame  $i $ to their corresponding locations in frame  $j $, as estimated RAFT~\cite{teed2020raft}. The induced flow  $\omega_{ij}(T_{ij}, d_i) $ is computed based on the current camera pose  $T_{ij}$ and depth map  $d_i$, and it represents the reprojection of pixels from frame  $i $ to frame  $j $:

\begin{equation}
    \omega_{ij}(T_{ij}, d_i) = \Pi_c(T_{ij} \circ \Pi_c^{-1}(p_i, d_i))
\end{equation}

\noindent
Here,  $\circ $ denotes function composition, where  $\Pi_c^{-1}(p_i, d_i) $ maps 2D pixel coordinates  $p_i \in \mathbb{R}^{H \times W \times 2} $ and the corresponding depth map  $d_i \in \mathbb{R}^{H \times W} $ to a 3D point cloud.
where $T_{ij} \in \text{SE}(3) $ is the relative pose transformation between frames $i $ and $j $, $\Pi_c $ is the camera projection operator, and $\Pi_c^{-1} $ is its inverse that maps 2D pixel coordinates $p_i $ and depth $d_i $ to a 3D point cloud.

The optimization minimizes the discrepancy between the predicted flow and the corrected induced flow, weighted by the confidence matrix:



\begin{equation}
    T, d = \min_{T, d} \sum_{(i,j) \in E} \left\| f^{\text{pred}}_{ij} - \left( \omega_{ij}(T_{ij}, d_i) + r_{ij} \right) \right\|^2_{\Sigma_{ij}}
\end{equation}

\noindent
where $r_{ij} $ is the flow correction term predicted by the ConvGRU, and $\Sigma_{ij} = \text{diag}(W_{ij}) $ is a diagonal confidence matrix derived during optimization.

To solve this optimization problem, we follow the Gauss-Newton algorithm implemented in DROID-SLAM~\cite{teed2021droid}. The algorithm linearizes the objective function and computes updates for  $\Delta \xi $ and  $\Delta d $ efficiently using the Schur complement. At each iteration, the updated camera poses  $T $ and depth maps  $d $ progressively refine the reconstruction. Detailed derivations and implementation steps are described in~\cite{teed2021droid}.

\subsection{Mapping}
\label{mapping}

Once frames data enters the system, it is first filtered through a \textit{motion filter} and a \textit{keyframe threshold} to determine whether the current frame qualifies as a keyframe. Keyframes are selected based on camera motion and frame characteristics and stored in a dynamic \textit{factor graph} $ G  $ as mentioned in \ref{sec:Graph}.

\subsubsection{Initialization of Factor Graph}\label{sec:graph}
During the initialization phase, initial frames are used to initialize the factor graph by establishing sufficient keyframes and constraints, providing a stable foundation for subsequent optimization. For each keyframe, we obtain the estimated camera pose and disparity maps, derived as $ 1/d  $ from the estimated depth $ d  $, via SLAM’s frontend tracking. Using these disparity maps, we then back-project 2D image points into 3D space, creating a dense sfm point cloud. This point cloud forms the basis of our scene model initialization.

\subsubsection{Gaussian Scene Modeling from Dense SLAM Point Cloud}\label{sec:3dgs1}
To model the dense SLAM point cloud, we represent the geometry as a set of 3D Gaussians, which capture the local geometric structure and uncertainty through covariance matrices. This transformation enhances the robustness of scene representation, addressing limitations of SLAM-generated point clouds, such as noise and lack of local continuity, and facilitates downstream optimization and rendering tasks within the 3DGS framework. Following the 3DGS approach, each Gaussian is defined by a full 3D covariance matrix $ \Sigma  $ in world space, with the mean $ \mu  $ located at each point:
\begin{equation}
\mathcal{G}(g) = \exp\left(-\frac{1}{2} (\bm{x} - \bm{\mu})^T \Sigma^{-1} (\bm{x} - \bm{\mu})\right)
\end{equation}
Where Gaussian is centered at $\mu$ and $ x $ is the position of a pixel. This Gaussian distribution is used to represent each point in the scene, capturing both its position and the uncertainty associated with it.

\subsubsection{Covariance Matrix Decomposition}\label{sec:3dgs2}
The covariance matrix $ \Sigma  $ of the 3D Gaussian distribution can be decomposed into a \textit{scaling matrix} $ S  $ and a \textit{rotation matrix} $ R  $, providing an intuitive yet powerful representation for optimization:
\begin{equation}
\Sigma = R S S^T R^T
\end{equation}
where $ S  $ is a diagonal matrix that encodes the scaling along different axes of the ellipsoid, and $ R  $ is a rotation matrix that describes the orientation of the ellipsoid in space. For practical purposes, we store $ S  $ as a 3D vector and $ R  $ as a quaternion, allowing independent optimization of both factors. This decomposition ensures efficient representation and facilitates optimization, as each element can be treated separately during gradient descent.

\subsubsection{SLAM-Informed Adaptive Densification}\label{sec:siad}

\begin{figure}[ht]
    \centering
    \includegraphics[width=0.6\linewidth]{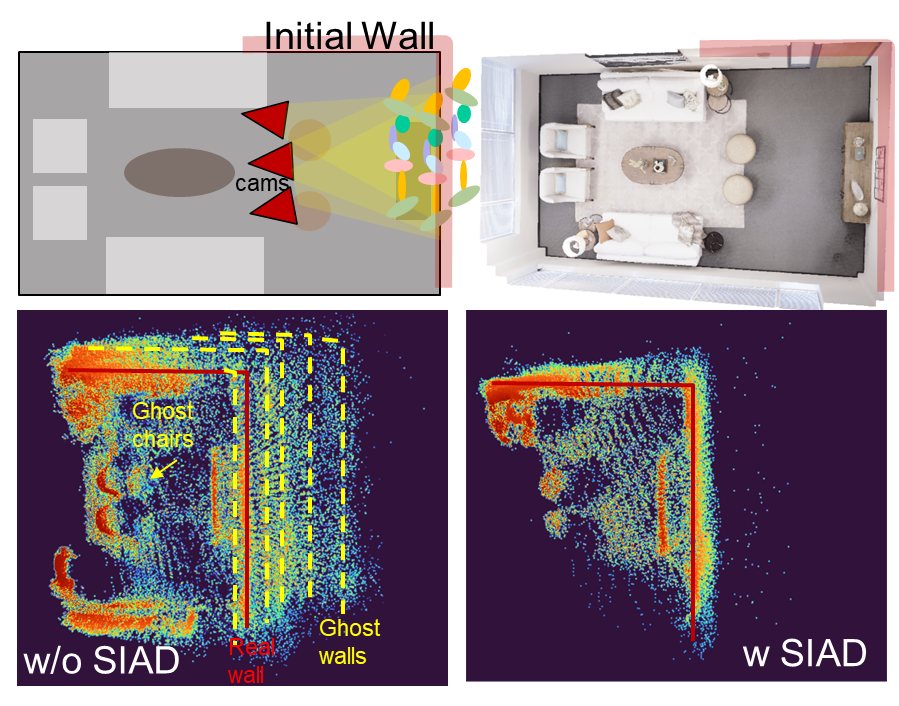}
    \caption{When the Gaussian scale is set to 0.01, the inaccuracy of Gaussian position estimation becomes evident, particularly during the initial phase of SLAM when data is sparse. The accumulation of non-updated Gaussian point clouds results in ghosting artifacts on walls and furniture, which degrades both the geometric quality of the reconstruction and the final rendering performance.}
    \label{fig:siad}
\end{figure}

Monocular SLAM systems often generate inaccurate point clouds during the early stages of mapping due to limited observations and uncertainties in depth and pose estimation. These inaccuracies arise from insufficient baseline for triangulation and weak pose constraints, which are particularly pronounced in the initialization phase. Such erroneous estimates can result in geometric and topological inconsistencies, inaccurate depth estimation, and error propagation to downstream rendering tasks. As shown in Fig.\ref{fig:siad} the Gaussians with wrong estimated depth will result in ``ghost" artifacts on walls and furniture, particularly when viewed from untrained perspectives. Moreover, these issues increase computational costs for Gaussian modeling, as additional resources are required to correct the initial inaccuracies.

While traditional 3DGS densification methods excel at densifying sparse point clouds (e.g., Photo-SLAM), they are inefficient for our system, which generates dense point clouds for each frame. Such methods fail to address the early-stage inaccuracies in SLAM-generated point clouds and add unnecessary computational overhead for densification. To overcome these challenges, we propose SLAM-Informed Adaptive Densification (SIAD), which leverages SLAM’s dynamic updates to refine the point cloud online. By pruning erroneous points, incorporating new ones, and adapting Gaussian representations based on reliable masks, our approach ensures accurate and efficient point cloud refinement without redundant densification steps. Our method circumvents the need for traditional cloning and splitting operations typically used in 3DGS densification, thereby streamlining the process and avoiding the associated computational overhead.

\paragraph{Point Reliability Mask Generation}\label{sec:mask}

Reliability masks in SLAM are critical for determining which points are valid or reliable in the dense depth and pose estimation process. In DROID-SLAM, these masks are derived and updated iteratively during tracking and optimization. 
Specifically, in our method the reliability masks $m_i$ are created based on the following consistency metrics:

\aptLtoX[graphic=no,type=html]{\begin{enumerate}
\item[(1)] \textbf{Depth Consistency Check} Each frame $ i  $ maintains a dense depth map $ d_i  $, and the corresponding mask $ m_i  $ is generated to indicate valid depth points:
\begin{equation}
    m_i(p) = 
    \begin{cases} 
    1, & \text{if } d_i(p) > 0 \text{ and } d_i(p) \text{ is consistent} \\
    0, & \text{otherwise}
    \end{cases}
\end{equation}
Here, $ p  $ represents a pixel in the frame. Validity is determined by the network's depth predictions and their geometric consistency across frames. A depth value $d_i(p)$ is considered consistent if it aligns geometrically with its reprojected counterpart in neighboring frames and remains stable across consecutive frames.

\item[(2)] \textbf{Frame-to-Frame Geometric Consistency} Our system uses camera poses $ T_{ij}  $ and depth maps $ d_i  $ to establish geometric correspondences between frames $i$ and $j$. Re-projection and distance calculations were defined as:
\begin{equation}
    \mathbf{x}_j = T_{ij} \cdot \mathbf{x}_i, \quad d(\mathbf{x}_i, \mathbf{x}_j) = \| \mathbf{x}_j - \hat{\mathbf{x}}_j \|
\end{equation}
where $ \mathbf{x}_i  $ and $ \mathbf{x}_j  $ are 3D points in frames $ i  $ and $ j  $, and $ \hat{\mathbf{x}}_j  $ is the reprojected point from $ \mathbf{x}_i  $. If $ d(\mathbf{x}_i, \mathbf{x}_j) > \text{thresh}  $, the point is marked invalid in $ m_i  $.

\item[(3)] \textbf{Factor Graph Confidence Weight} During optimization, reliable edges are those that exhibit higher weights $ w_{ij}  $, indicating stronger geometric consistency between the connected frames. Such edges are preserved in the graph, and the associated points in the masks $ m_i  $ of the connected frames are retained. The weight $ w_{ij}  $ reflects the confidence of the correspondence between the two frames, based on the projection and matching quality of their respective points.

For individual points, their weights $ w_{ij}(p) $ are evaluated independently from the conv-GRU. Points with low confidence weights, where $ w_{ij}(p) < \epsilon  $, are considered unreliable and are removed from the graph. This removal is reflected in the mask $ m_i  $ of the corresponding frame, with the invalidated points marked as:
\begin{equation}
    m_i(p) = 0, \quad \text{if } \text{mean}(w_{ij}(p)) < \epsilon
\end{equation}
\end{enumerate}}{\begin{enumerate}[leftmargin=1cm, itemindent=\parindent, noitemsep, label=(\arabic*)]
\item \textbf{Depth Consistency Check} Each frame $ i  $ maintains a dense depth map $ d_i  $, and the corresponding mask $ m_i  $ is generated to indicate valid depth points:
\begin{equation}
    m_i(p) = 
    \begin{cases} 
    1, & \text{if } d_i(p) > 0 \text{ and } d_i(p) \text{ is consistent} \\
    0, & \text{otherwise}
    \end{cases}
\end{equation}
Here, $ p  $ represents a pixel in the frame. Validity is determined by the network's depth predictions and their geometric consistency across frames. A depth value $d_i(p)$ is considered consistent if it aligns geometrically with its reprojected counterpart in neighboring frames and remains stable across consecutive frames.

\item \textbf{Frame-to-Frame Geometric Consistency} Our system uses camera poses $ T_{ij}  $ and depth maps $ d_i  $ to establish geometric correspondences between frames $i$ and $j$. Re-projection and distance calculations were defined as:
\begin{equation}
    \mathbf{x}_j = T_{ij} \cdot \mathbf{x}_i, \quad d(\mathbf{x}_i, \mathbf{x}_j) = \| \mathbf{x}_j - \hat{\mathbf{x}}_j \|
\end{equation}
where $ \mathbf{x}_i  $ and $ \mathbf{x}_j  $ are 3D points in frames $ i  $ and $ j  $, and $ \hat{\mathbf{x}}_j  $ is the reprojected point from $ \mathbf{x}_i  $. If $ d(\mathbf{x}_i, \mathbf{x}_j) > \text{thresh}  $, the point is marked invalid in $ m_i  $.

\item \textbf{Factor Graph Confidence Weight} During optimization, reliable edges are those that exhibit higher weights $ w_{ij}  $, indicating stronger geometric consistency between the connected frames. Such edges are preserved in the graph, and the associated points in the masks $ m_i  $ of the connected frames are retained. The weight $ w_{ij}  $ reflects the confidence of the correspondence between the two frames, based on the projection and matching quality of their respective points.

For individual points, their weights $ w_{ij}(p) $ are evaluated independently from the conv-GRU. Points with low confidence weights, where $ w_{ij}(p) < \epsilon  $, are considered unreliable and are removed from the graph. This removal is reflected in the mask $ m_i  $ of the corresponding frame, with the invalidated points marked as:
\begin{equation}
    m_i(p) = 0, \quad \text{if } \text{mean}(w_{ij}(p)) < \epsilon
\end{equation}
\end{enumerate}}

\paragraph{Dynamic Mask Updates}\label{sec:maskupdate}
Reliability masks are updated dynamically by evaluating geometric distances between frames, and points are marked valid or invalid based on proximity thresholds. This ensures that masks reflect changes in scene structure and camera trajectory during optimization.

\paragraph{SLAM-Informed Adaptive Densification}

The SLAM-Informed Adaptive Densification process leverages the dynamically updated masks to adaptively adjust the point cloud's density, ensuring it reflects the current scene structure accurately. This process consists of three core components:

\aptLtoX[graphic=no,type=html]{\begin{enumerate}
\item[(1)] \textbf{Position Updates for Existing Points} For points that remain valid in the updated masks, 
$ m^{(k-1)}(p) = 1  $ and $ m^{(k)}(p) = 1  $, their positions are updated based on SLAM's tracking results. Given the incremental camera pose update $ \Delta \xi ^{(k)}  $, the updated position $ g_p^{(k)}  $ of a point $ g_p^{(k-1)}  $ is:
\begin{equation}
\label{equ:uodate_loc}
    g_p^{(k)} = \text{exp}(\Delta \xi^{(k)})\circ T_{ij} \circ \Pi_c^{-1}\big(p, d_i^{(k-1)} + \Delta d_i^{(k)}\big)
\end{equation}

\item[(2)] \textbf{Pruning of Invalid Points} For points that are marked invalid in the updated masks, 
$ m^{(k-1)}(p) = 1  $ and $ m^{(k)}(p) = 0 $, are pruned from the point cloud:
\begin{equation}
    g_p^{(k)} \rightarrow \varnothing, \quad \text{if } m^{(k)}(p) = 0
\end{equation}
This ensures that outdated or unreliable points, as determined by SLAM, do not pollute the Gaussian splatting representation.

\item[(3)] \textbf{Densification for Newly Valid Points} For points that are newly added to the mask, $ m^{(k-1)}(p) = 0  $ and $ m^{(k)}(p) = 1  $, we generated new Gaussian splatting points by initializing the points location using Eq.\ref{equ:uodate_loc}.
\end{enumerate}}{\begin{enumerate}[leftmargin=1cm,itemindent=2\parindent,noitemsep,label=(\arabic*)]
\item \textbf{Position Updates for Existing Points} For points that remain valid in the updated masks, 
$ m^{(k-1)}(p) = 1  $ and $ m^{(k)}(p) = 1  $, their positions are updated based on SLAM's tracking results. Given the incremental camera pose update $ \Delta \xi ^{(k)}  $, the updated position $ g_p^{(k)}  $ of a point $ g_p^{(k-1)}  $ is:
\begin{equation}
\label{equ:uodate_loc}
    g_p^{(k)} = \text{exp}(\Delta \xi^{(k)})\circ T_{ij} \circ \Pi_c^{-1}\big(p, d_i^{(k-1)} + \Delta d_i^{(k)}\big)
\end{equation}

\item \textbf{Pruning of Invalid Points} For points that are marked invalid in the updated masks, 
$ m^{(k-1)}(p) = 1  $ and $ m^{(k)}(p) = 0 $, are pruned from the point cloud:
\begin{equation}
    g_p^{(k)} \rightarrow \varnothing, \quad \text{if } m^{(k)}(p) = 0
\end{equation}
This ensures that outdated or unreliable points, as determined by SLAM, do not pollute the Gaussian splatting representation.

\item \textbf{Densification for Newly Valid Points} For points that are newly added to the mask, $ m^{(k-1)}(p) = 0  $ and $ m^{(k)}(p) = 1  $, we generated new Gaussian splatting points by initializing the points location using Eq.\ref{equ:uodate_loc}.
\end{enumerate}}

\subsubsection{Keyframe Selection and Optimization Strategy}\label{sec:keyframe}

To efficiently handle the temporal sequence of SLAM and maintain geometric accuracy during mapping, our system continuously processes incoming video frames in the SLAM frontend. A motion-based filter, inspired by~\cite{teed2021droid}, is employed to identify and select keyframes. This filter evaluates the relative motion between frames, prioritizing frames with significant pose differences or high optical flow magnitudes, while discarding redundant frames. The selected keyframes ensure that the SLAM system captures sufficient geometric diversity for accurate reconstruction without overwhelming computational resources.

We maintain a frontend keyframe window with a default size of 25. This window size is chosen empirically to balance the
need for capturing sufficient geometric diversity and controlling computational overhead. Our
preliminary experiments indicated that a window size of 25 is effective in maintaining a robust
optimization process while limiting the complexity of the bundle adjustment-like updates.
Similar strategies have been adopted in related work, where keyframe retention is controlled
via a maximum age parameter to ensure a balance between temporal context and near real-time
performance. At each iteration, the system updates the camera poses and depth estimates for all keyframes within this window using a bundle adjustment-like optimization process. These updates are subsequently propagated to the mapping module, where the dense point cloud is refined and reconstructed into a Gaussian-based scene representation.

\paragraph{Mapping with 3D Gaussian Splatting}
In the mapping module, the SLAM outputs—updated camera poses and depth maps—are used to refine the 3D Gaussian scene representation. The means $\mu$ of the Gaussians, representing point positions, are adjusted based on the updated depth and pose information. Specifically, the updated depth maps provide refined 3D coordinates, while pose updates ensure consistent alignment within the global frame. Unlike the means, the covariance matrices $\Sigma$, which define the shape and uncertainty of each Gaussian, are optimized using downstream photometric and geometric constraints rather than directly by SLAM outputs.

We adopt the same rendering method with 3DGS, in which each Gaussian is projected onto the image plane as a 2D elliptical footprint based on its mean $\mu$ and covariance matrix $\Sigma$. The rendering pipeline accumulates contributions from all Gaussians through a rasterization process, where the opacity and color of each pixel are computed by blending overlapping Gaussian footprints. This method ensures efficient rendering while preserving high visual fidelity and geometric accuracy, making it particularly suitable for online applications.

\paragraph{Geometry-Guided Optimization}
To achieve high-quality rendering that balances photometric consistency and geometric accuracy, we define a composite loss function. The RGB loss $ L_{\text{rgb}} $, inspired by Plenoxels \cite{yu2021plenoxels} and 3DGS, ensures consistency between the rendered and ground truth images. Unlike 3DGS, which uses a combination of $L_1$ loss and SSIM loss, we enhance perceptual quality by incorporating multi-scale SSIM (MS-SSIM)~\cite{wang2003multiscale}, which captures perceptual consistency across different resolutions. Specifically, MS-SSIM is computed by first calculating the SSIM between the rendered image and the ground truth using a fixed window size of 11. Then, the input images are progressively downsampled via 2D average pooling with a stride of 2, and the SSIM is re-evaluated at each of the three scales. The final MS-SSIM loss is obtained by averaging the SSIM values across these scales. The RGB loss is defined as:

\begin{equation}
L_{\text{rgb}} = (1 - \lambda_{\text{ms-ssim}}) L_1 + \lambda_{\text{ms-ssim}} L_{\text{ms-ssim}}
\end{equation}

Here $ L_1  $ minimizes pixel-wise differences, and $ L_{\text{ms-ssim}}  $  improves perceptual quality across multiple scales. This multi-scale approach ensures robustness to local variations and yields better perceptual alignment compared to the single-scale SSIM used in 3DGS.

To further capture the geometric relationships within the scene, we add an edge-aware normal loss $ L_{\text{normal}}  $~\cite{liu2024atomgs} into our system. This loss uses the normal map $ \mathbf{N}  $, computed from the gradients of the depth map $ \mathbf{D}  $ as:

\begin{equation}
    \mathbf{N} = \frac{\nabla_x \bm{D} \times \nabla_y \bm{D}}{\|\nabla_x \bm{D} \times \nabla_y \bm{D}\|}
\end{equation}

\noindent
and penalizes the gradient of $ \mathbf{N}  $ modulated by a weight function $ \omega(x) = (x - 1)^q  $, which balances smoothing in low-gradient areas and detail preservation at sharp edges:
\begin{equation}
L_{\text{geo}} = \frac{1}{HW} \sum_{i}^H \sum_{j}^W \left| \nabla \bm{N} \right| \otimes \omega(|\nabla \bm{I}|)
\end{equation}

We noticed that adding the edge-aware loss is effective for emphasizing geometric details but it amplifies the gradient penalties in high-gradient regions and suppresses them in flat areas. While effective for emphasizing geometric details, this approach may overemphasize sharp edges, resulting in very limited improvements in RGB rendering quality, reducing overall scene consistency.
To address this, we propose a smooth weighting function that balances edge emphasis and spatial continuity using a Gaussian-like modulation:

\begin{equation}\label{equ:smooth}
    \omega(x) = \exp\left(-\frac{|x - 1|^2}{\sigma^2}\right)
\end{equation}

\noindent
where \(\sigma\) controls the smoothness of the weighting, ensuring gradual transitions between high- and low-gradient regions. This modification reduces the over-penalization of sharp gradients and prevents excessively low penalties in flat areas, improving RGB rendering consistency without sacrificing geometric accuracy.

The final composite loss function integrates the RGB loss, the geometry-guided edge-aware normal loss is calculated:
\begin{equation}
L = L_{\text{rgb}} + \lambda_{\text{geo}} L_{\text{geo}} 
\end{equation}

\section{Evaluation}
\subsection{Implementation and Experiment Setup}

We evaluated SplatMap against several state-of-the-art (SOTA) SLAM systems, focusing on both RGB-D and monocular setups. For RGB-D systems, we compared with the neural radiance field (NeRF)-based SLAM systems \textit{NICE-SLAM}\cite{zhu2022nice} and \textit{ESLAM}, as well as Gaussian-based SLAM methods including \textit{GS-SLAM}\cite{yan2024gs} and \textit{SplaTAM}\cite{keetha2024splatam}. For monocular systems, we selected \textit{GLORIE-SLAM}\cite{zhang2024glorie} and \textit{Photo-SLAM}~\cite{huang2024photo}.

We evaluated SplatMAP on the \textit{Replica} dataset \cite{straub2019replicadatasetdigitalreplica} and the \textit{TUM RGB-D} dataset \cite{fisher2021colmap}. We used the Absolute Trajectory Error (ATE) metric \cite{fisher2021colmap} to assess localization accuracy, reporting both the RMSE and STD of ATE. Quantitative evaluations of photorealistic mapping performance were conducted using PSNR, SSIM, and LPIPS metrics.  

All comparison methods using their official implementations, SplatMap and the comparison methods were run on a desktop equipped with an NVIDIA RTX 4090 24 GB GPU, an Intel Core i9-13900K CPU, and 64 GB of RAM, and A6000 Server with 50GB GPU.

\subsection{Tracking Results}

We evaluated several state-of-the-art RGB and RGB-D SLAM methods on the Replica dataset using the root mean square error (RMSE) of Absolute Trajectory Error (ATE), as shown in Table~\ref{tab:ATE_replica}. Traditional methods like NICE-SLAM and Vox-Fusion exhibited higher ATE values (1.060 and 3.090, respectively), reflecting challenges in accurate trajectory estimation. Advanced methods such as SplaTAM (0.360), GS-SLAM (0.500), and Gaussian Splatting SLAM (0.580) demonstrated improved performance, with monocular systems like GLORIE-SLAM achieving a competitive ATE of 0.310.

Our proposed method achieved the best overall performance, with an average ATE of 0.179, a 50\% reduction compared to SplaTAM. Notably, it recorded the lowest ATE in five out of nine scenes, including Room0 (0.040), Room1 (0.190), and Office0 (0.094). These results highlight the superior accuracy and robustness of our approach in trajectory estimation, providing a reliable foundation for the following mapping and rendering tasks.

\begin{table*}[!t]
    \centering
    \caption{Average Trajectory Error (ATE) for Different SLAM Methods.}
\resizebox{\textwidth}{!}{
    \begin{tabular}{lccccccccc}
        \toprule
         \textbf{Method} & \textbf{AVE} & \textbf{Room0} & \textbf{Room1} & \textbf{Room2} & \textbf{Office0} & \textbf{Office1} & \textbf{Office2} & \textbf{Office3} & \textbf{Office4} \\
        \midrule
        NICE-SLAM~\cite{zhu2022nice}    & 1.06          & 0.23          & 1.31          & 1.07          & 1.00          & 1.06         & 1.10          & 1.13          & 1.06          \\
        Vox-Fusion~\cite{yang2022vox}   & 3.09         & 0.24          & 4.70          & 1.47          & 8.48         & 2.04        & 2.58          & 1.110          & 2.94          \\
        ESLAM~\cite{johari2023eslamefficientdenseslam}        & 0.63          & 0.25        & 0.70          & 0.52          & 0.57          & 0.55          & 0.58          & 0.72          & 0.63 \\
        Point-SLAM~\cite{sandstrom2023point}   & 0.52          & 0.12          & 0.41         & 0.37         & 0.38 & 0.48         & 0.54          & 0.69          & 0.72          \\
        GS Slam~\cite{yan2024gsslamdensevisualslam}      & 0.50         & 0.08          & 0.53          & 0.33          & 0.53          & 0.41          & 0.46          & 0.70          & 0.70          \\
        SplaTAM~\cite{keetha2024splatam}      & \cellcolor{yellow!25}0.36 & 0.10          & 0.40          & 0.29          & 0.47          & \cellcolor{orange!25}0.27  & \cellcolor{orange!25}0.29 & 0.55          & 0.72          \\
        MonoGS~\cite{matsuki2024gaussian}       & 0.58          & \cellcolor{orange!25}0.07 & 0.37 & \cellcolor{yellow!25}0.23 & 0.66          & 0.72          & \cellcolor{yellow!25}0.30 & \cellcolor{red!25}0.19  & 1.46          \\
        GLORIE-SLAM~\cite{zhang2024glorie}  & \cellcolor{orange!25}0.31 & 0.52          & \cellcolor{yellow!25}0.21 & 0.26          & \cellcolor{yellow!25}0.29 & 0.41          & 0.46          & 0.44 & \cellcolor{yellow!25}0.36 \\
        Photo-SLAM~\cite{huang2024photo}   & 1.09         & \cellcolor{yellow!25}0.08 & 1.18         & \cellcolor{orange!25}0.23 & 0.58          & 0.32 & 0.69          & 0.72          & 0.72          \\
        RTG-slam~\cite{peng2024rtg}   & \cellcolor{red!25}0.18         & 0.20 & \cellcolor{red!25}0.18        & \cellcolor{red!25}0.13 & \cellcolor{orange!25}0.22          & \cellcolor{red!25}0.12 & \cellcolor{orange!25}0.22          &  \cellcolor{orange!25}0.20          & \cellcolor{orange!25}0.19          \\
        Ours         & \cellcolor{red!25}0.18 & \cellcolor{red!25}0.04 & \cellcolor{orange!25}0.19 & \cellcolor{red!25}0.13& \cellcolor{red!25}0.09 & \cellcolor{yellow!25}0.31 & \cellcolor{red!25}0.17 & \cellcolor{yellow!25}0.32 & \cellcolor{red!25}0.06 \\
        \bottomrule
    \end{tabular}}
    \label{tab:ATE_replica}
\end{table*}

\subsection{Mapping Results}
We evaluated the performance of SplatMap on the Replica and TUM-RGBD datasets using PSNR, SSIM, and LPIPS metrics, as shown in Tab.~\ref{tab:map_replica} and~\ref{tab:map_tum}. While our system is monocular, it outperformed not only state-of-the-art (SOTA) monocular SLAM systems but also surpassed several RGB-D SLAM systems, demonstrating the robustness and accuracy of our approach.

\aptLtoX[graphic=no,type=html]{\begin{table*}[!t]
\centering
\caption{PSNR, SSIM, and LPIPS Metrics for Different SLAM Methods on Replica Dataset.}
\resizebox{\textwidth}{!}{
\begin{tabular}{|c|c|c|c|c|c|c|c|c|c|c|c|}
\toprule
 Input & Method & Metric 
 & AVE & Room0 & Room1 & Room2 & Office0 & Office1 & Office2 & Office3 & Office4 \\
\midrule
\multirow{22}{*}{RGBD} 
& \multirow{3}{*}{NICE-SLAM~\cite{zhu2022nice}} 
 & PSNR  
   & 24.42
   & 22.12
   & 22.47
   & 24.52
   & 29.07
   & 30.34
   & 19.66
   & 22.23
   & 24.49 \\
 &  & SSIM 
   & 0.81  
   & 0.69  
   & 0.76  
   & 0.81  
   & 0.87
   & 0.89
   & 0.80  
   & 0.80  
   & 0.86 \\
 &  & LPIPS 
   & 0.23
   & 0.33
   & 0.27
   & 0.21
   & 0.230
   & 0.18
   & 0.24
   & 0.21
   & 0.200 \\
\cline{2-12}
& \multirow{3}{*}{Vox-Fusion~\cite{yang2022vox}} 
 & PSNR  
   & 24.41
   & 22.39
   & 22.36
   & 23.92
   & 27.79
   & 29.83
   & 20.33
   & 23.47
   & 25.21 \\
 &  & SSIM 
   & 0.80
   & 0.68
   & 0.68
   & 0.80
   & 0.86
   & 0.88
   & 0.79
   & 0.80
   & 0.85 \\
 &  & LPIPS 
   & 0.24
   & 0.30
   & 0.30
   & 0.23
   & 0.24
   & 0.18
   & 0.24
   & 0.21
   & 0.20 \\
\cline{2-12}

& \multirow{3}{*}{ESLAM~\cite{johari2023eslamefficientdenseslam}} 
 & PSNR  
   & 29.08
   & 25.32
   & 27.77
   & 29.08
   & 33.71
   & 30.20
   & 28.09
   & 28.77
   & 29.71 \\
 &  & SSIM 
   & 0.93
   & 0.88
   & 0.90
   & 0.93
   & 0.96
   & 0.92
   & 0.94
   & 0.95
   & 0.95 \\
 &  & LPIPS 
   & 0.25
   & 0.31
   & 0.30
   & 0.25
   & 0.18
   & 0.23
   & 0.24
   & 0.20
   & 0.200 \\
\cline{2-12}

& \multirow{3}{*}{Point-SLAM~\cite{sandstrom2023point}} 
 & PSNR  
   & \cellcolor{orange!25}35.17
   & \cellcolor{yellow!25}32.40
   & \cellcolor{red!25}34.08
   & \cellcolor{orange!25}35.50
   & \cellcolor{yellow!25}38.26
   & 39.16
   & \cellcolor{orange!25}33.99
   & \cellcolor{orange!25}33.48
   & \cellcolor{yellow!25}33.49 \\
 &  & SSIM 
   & \cellcolor{orange!25}0.98  
   & \cellcolor{yellow!25}0.97  
   & \cellcolor{red!25}0.98     
   & \cellcolor{orange!25}0.98  
   & 0.98                       
   & 0.99                       
   & 0.96                       
   & 0.96                       
   & 0.98                       
   \\
 &  & LPIPS 
   & 0.12
   & 0.11
   & 0.12
   & 0.11
   & 0.10
   & 0.12
   & 0.16
   & 0.13
   & 0.14 \\
\cline{2-12}

& \multirow{3}{*}{GS Slam~\cite{yan2024gsslamdensevisualslam}} 
 & PSNR  
   & \cellcolor{yellow!25}34.27
   & 31.56
   & \cellcolor{yellow!25}32.86
   & 32.56
   & \cellcolor{orange!25}38.70
   & \cellcolor{orange!25}41.17
   & \cellcolor{yellow!25}32.36
   & \cellcolor{yellow!25}32.03
   & 32.92 \\
 &  & SSIM 
   & \cellcolor{yellow!25}0.98
   & 0.97
   & \cellcolor{orange!25}0.97
   & 0.97
   & \cellcolor{orange!25}0.99
   & \cellcolor{red!25}0.99
   & 0.98
   & 0.97
   & 0.97 \\
 &  & LPIPS 
   & \cellcolor{yellow!25}0.08
   & 0.09
   & \cellcolor{yellow!25}0.08
   & 0.09
   & \cellcolor{yellow!25}0.05
   & \cellcolor{red!25}0.03
   & 0.09
   & 0.11
   & 0.11 \\
\cline{2-12}
& \multirow{3}{*}{SplaTAM~\cite{keetha2024splatam} }
 & PSNR  
   & 34.110          
   & \cellcolor{orange!25}32.86  
   & \cellcolor{orange!25}33.89  
   & \cellcolor{yellow!25}35.25  
   & \cellcolor{yellow!25}38.26  
   & \cellcolor{yellow!25}39.17  
   & 31.97          
   & 29.70          
   & 31.81          
   \\
 &  & SSIM
   & 0.97           
   & \cellcolor{orange!25}0.98  
   & \cellcolor{yellow!25}0.97  
   & \cellcolor{orange!25}0.98  
   & \cellcolor{yellow!25}0.98  
   & 0.97           
   & \cellcolor{yellow!25}0.97  
   & 0.95           
   & 0.95           
   \\
 &  & LPIPS
   & 0.10           
   & \cellcolor{orange!25}0.07  
   & 0.10           
   & 0.08           
   & 0.09           
   & 0.10           
   & 0.10           
   & 0.12           
   & 0.15           
   \\
\cline{2-12}
& \multirow{3}{*}{RTG-SLAM~\cite{peng2024rtg}} 
 & PSNR  
   & 35.43
   & 31.56
   & 34.21
   & 35.57
   & 39.11
   & 40.27
   & 33.54
   & 32.76
   & 36.48 \\
 &  & SSIM 
   & 0.967  
   & 0.979  
   & 0.981  
   & 0.990  
   & 0.992
   & 0.981
   & 0.981  
   & 0.984  
   & 0.982 \\
 &  & LPIPS 
   & 0.131
   & 0.105
   & 0.115
   & 0.068
   & 0.075
   & 0.134
   & 0.128
   & 0.117
   & 0.109 \\
\cline{2-12}
\cline{1-12}
\multirow{9}{*}{MONO} 
& \multirow{3}{*}{GLORIE-SLAM~\cite{zhang2024glorie}}
 & PSNR
   & 31.04
   & 28.49
   & 30.09
   & 29.98
   & 35.88
   & 37.15
   & 28.45
   & 28.54
   & 29.73
   \\
 &  & SSIM
   & 0.97
   & 0.96
   & \cellcolor{yellow!25}0.97  
   & 0.96
   & 0.98
   & \cellcolor{yellow!25}0.99  
   & \cellcolor{yellow!25}0.97  
   & \cellcolor{orange!25}0.97  
   & \cellcolor{yellow!25}0.97  
   \\
 &  & LPIPS
   & 0.12
   & 0.13
   & 0.13
   & 0.14
   & 0.09
   & 0.08
   & 0.15
   & 0.11
   & 0.15
   \\
\cline{2-12}
& \multirow{3}{*}{PhotoSLAM~\cite{huang2024photo}}
 & PSNR 
   & 33.30
   & 29.77
   & 31.30
   & 33.18
   & 36.99
   & 37.59
   & 31.79
   & 31.62
   & \cellcolor{red!25}34.17  
   \\
 &  & SSIM 
   & 0.93
   & 0.87
   & 0.91
   & 0.93
   & 0.96
   & 0.95
   & 0.93
   & 0.92
   & 0.94
   \\
 &  & LPIPS 
   & \cellcolor{orange!25}0.079  
   & 0.11
   & 0.08
   & 0.07
   & 0.06
   & 0.06
   & 0.09
   & 0.09
   & 0.07
   \\
\cline{2-12}
& \multirow{3}{*}{Ours}
 & PSNR  
   & \cellcolor{red!25}36.864  
   & \cellcolor{red!25}35.367  
   & 31.746
   & \cellcolor{red!25}38.111  
   & \cellcolor{red!25}42.888  
   & \cellcolor{red!25}42.082  
   & \cellcolor{red!25}35.785  
   & \cellcolor{red!25}35.004  
   & \cellcolor{orange!25}33.930  
   \\
 &  & SSIM 
   & \cellcolor{red!25}0.985  
   & \cellcolor{red!25}0.988  
   & 0.950
   & \cellcolor{red!25}0.992  
   & \cellcolor{red!25}0.996  
   & \cellcolor{red!25}0.995  
   & \cellcolor{red!25}0.990  
   & \cellcolor{red!25}0.991  
   & \cellcolor{orange!25}0.974  
   \\
 &  & LPIPS 
   & \cellcolor{red!25}0.040  
   & \cellcolor{red!25}0.031  
   & 0.088
   & \cellcolor{red!25}0.026  
   & \cellcolor{red!25}0.014  
   & \cellcolor{red!25}0.019  
   & \cellcolor{red!25}0.039  
   & \cellcolor{red!25}0.039  
   & 0.064
   \\
\cline{1-12}
\end{tabular}}
\label{tab:map_replica}
\end{table*}}{\begin{table*}[!t]
\centering
\caption{PSNR, SSIM, and LPIPS Metrics for Different SLAM Methods on Replica Dataset.}
\resizebox{\textwidth}{!}{
\begin{tabular}{|c|c|c|c|c|c|c|c|c|c|c|c|}
\hline
 Input & Method & Metric 
 & AVE & Room0 & Room1 & Room2 & Office0 & Office1 & Office2 & Office3 & Office4 \\
\hline
\multirow{22}{*}{RGBD} 
& \multirow{3}{*}{NICE-SLAM~\cite{zhu2022nice}} 
 & PSNR  
   & 24.42
   & 22.12
   & 22.47
   & 24.52
   & 29.07
   & 30.34
   & 19.66
   & 22.23
   & 24.49 \\
 &  & SSIM 
   & 0.81  
   & 0.69  
   & 0.76  
   & 0.81  
   & 0.87
   & 0.89
   & 0.80  
   & 0.80  
   & 0.86 \\
 &  & LPIPS 
   & 0.23
   & 0.33
   & 0.27
   & 0.21
   & 0.230
   & 0.18
   & 0.24
   & 0.21
   & 0.200 \\
\cline{2-12}
& \multirow{3}{*}{Vox-Fusion~\cite{yang2022vox}} 
 & PSNR  
   & 24.41
   & 22.39
   & 22.36
   & 23.92
   & 27.79
   & 29.83
   & 20.33
   & 23.47
   & 25.21 \\
 &  & SSIM 
   & 0.80
   & 0.68
   & 0.68
   & 0.80
   & 0.86
   & 0.88
   & 0.79
   & 0.80
   & 0.85 \\
 &  & LPIPS 
   & 0.24
   & 0.30
   & 0.30
   & 0.23
   & 0.24
   & 0.18
   & 0.24
   & 0.21
   & 0.20 \\
\cline{2-12}

& \multirow{3}{*}{ESLAM~\cite{johari2023eslamefficientdenseslam}} 
 & PSNR  
   & 29.08
   & 25.32
   & 27.77
   & 29.08
   & 33.71
   & 30.20
   & 28.09
   & 28.77
   & 29.71 \\
 &  & SSIM 
   & 0.93
   & 0.88
   & 0.90
   & 0.93
   & 0.96
   & 0.92
   & 0.94
   & 0.95
   & 0.95 \\
 &  & LPIPS 
   & 0.25
   & 0.31
   & 0.30
   & 0.25
   & 0.18
   & 0.23
   & 0.24
   & 0.20
   & 0.200 \\
\cline{2-12}

& \multirow{3}{*}{Point-SLAM~\cite{sandstrom2023point}} 
 & PSNR  
   & \cellcolor{orange!25}35.17
   & \cellcolor{yellow!25}32.40
   & \cellcolor{red!25}34.08
   & \cellcolor{orange!25}35.50
   & \cellcolor{yellow!25}38.26
   & 39.16
   & \cellcolor{orange!25}33.99
   & \cellcolor{orange!25}33.48
   & \cellcolor{yellow!25}33.49 \\
 &  & SSIM 
   & \cellcolor{orange!25}0.98  
   & \cellcolor{yellow!25}0.97  
   & \cellcolor{red!25}0.98     
   & \cellcolor{orange!25}0.98  
   & 0.98                       
   & 0.99                       
   & 0.96                       
   & 0.96                       
   & 0.98                       
   \\
 &  & LPIPS 
   & 0.12
   & 0.11
   & 0.12
   & 0.11
   & 0.10
   & 0.12
   & 0.16
   & 0.13
   & 0.14 \\
\cline{2-12}

& \multirow{3}{*}{GS Slam~\cite{yan2024gsslamdensevisualslam}} 
 & PSNR  
   & \cellcolor{yellow!25}34.27
   & 31.56
   & \cellcolor{yellow!25}32.86
   & 32.56
   & \cellcolor{orange!25}38.70
   & \cellcolor{orange!25}41.17
   & \cellcolor{yellow!25}32.36
   & \cellcolor{yellow!25}32.03
   & 32.92 \\
 &  & SSIM 
   & \cellcolor{yellow!25}0.98
   & 0.97
   & \cellcolor{orange!25}0.97
   & 0.97
   & \cellcolor{orange!25}0.99
   & \cellcolor{red!25}0.99
   & 0.98
   & 0.97
   & 0.97 \\
 &  & LPIPS 
   & \cellcolor{yellow!25}0.08
   & 0.09
   & \cellcolor{yellow!25}0.08
   & 0.09
   & \cellcolor{yellow!25}0.05
   & \cellcolor{red!25}0.03
   & 0.09
   & 0.11
   & 0.11 \\
\cline{2-12}
& \multirow{3}{*}{SplaTAM~\cite{keetha2024splatam} }
 & PSNR  
   & 34.110          
   & \cellcolor{orange!25}32.86  
   & \cellcolor{orange!25}33.89  
   & \cellcolor{yellow!25}35.25  
   & \cellcolor{yellow!25}38.26  
   & \cellcolor{yellow!25}39.17  
   & 31.97          
   & 29.70          
   & 31.81          
   \\
 &  & SSIM
   & 0.97           
   & \cellcolor{orange!25}0.98  
   & \cellcolor{yellow!25}0.97  
   & \cellcolor{orange!25}0.98  
   & \cellcolor{yellow!25}0.98  
   & 0.97           
   & \cellcolor{yellow!25}0.97  
   & 0.95           
   & 0.95           
   \\
 &  & LPIPS
   & 0.10           
   & \cellcolor{orange!25}0.07  
   & 0.10           
   & 0.08           
   & 0.09           
   & 0.10           
   & 0.10           
   & 0.12           
   & 0.15           
   \\
\cline{2-12}
& \multirow{3}{*}{RTG-SLAM~\cite{peng2024rtg}} 
 & PSNR  
   & 35.43
   & 31.56
   & 34.21
   & 35.57
   & 39.11
   & 40.27
   & 33.54
   & 32.76
   & 36.48 \\
 &  & SSIM 
   & 0.967  
   & 0.979  
   & 0.981  
   & 0.990  
   & 0.992
   & 0.981
   & 0.981  
   & 0.984  
   & 0.982 \\
 &  & LPIPS 
   & 0.131
   & 0.105
   & 0.115
   & 0.068
   & 0.075
   & 0.134
   & 0.128
   & 0.117
   & 0.109 \\
\cline{2-12}
\hline
\multirow{9}{*}{MONO} 
& \multirow{3}{*}{GLORIE-SLAM~\cite{zhang2024glorie}}
 & PSNR
   & 31.04
   & 28.49
   & 30.09
   & 29.98
   & 35.88
   & 37.15
   & 28.45
   & 28.54
   & 29.73
   \\
 &  & SSIM
   & 0.97
   & 0.96
   & \cellcolor{yellow!25}0.97  
   & 0.96
   & 0.98
   & \cellcolor{yellow!25}0.99  
   & \cellcolor{yellow!25}0.97  
   & \cellcolor{orange!25}0.97  
   & \cellcolor{yellow!25}0.97  
   \\
 &  & LPIPS
   & 0.12
   & 0.13
   & 0.13
   & 0.14
   & 0.09
   & 0.08
   & 0.15
   & 0.11
   & 0.15
   \\
\cline{2-12}
& \multirow{3}{*}{PhotoSLAM~\cite{huang2024photo}}
 & PSNR 
   & 33.30
   & 29.77
   & 31.30
   & 33.18
   & 36.99
   & 37.59
   & 31.79
   & 31.62
   & \cellcolor{red!25}34.17  
   \\
 &  & SSIM 
   & 0.93
   & 0.87
   & 0.91
   & 0.93
   & 0.96
   & 0.95
   & 0.93
   & 0.92
   & 0.94
   \\
 &  & LPIPS 
   & \cellcolor{orange!25}0.079  
   & 0.11
   & 0.08
   & 0.07
   & 0.06
   & 0.06
   & 0.09
   & 0.09
   & 0.07
   \\
\cline{2-12}
& \multirow{3}{*}{Ours}
 & PSNR  
   & \cellcolor{red!25}36.864  
   & \cellcolor{red!25}35.367  
   & 31.746
   & \cellcolor{red!25}38.111  
   & \cellcolor{red!25}42.888  
   & \cellcolor{red!25}42.082  
   & \cellcolor{red!25}35.785  
   & \cellcolor{red!25}35.004  
   & \cellcolor{orange!25}33.930  
   \\
 &  & SSIM 
   & \cellcolor{red!25}0.985  
   & \cellcolor{red!25}0.988  
   & 0.950
   & \cellcolor{red!25}0.992  
   & \cellcolor{red!25}0.996  
   & \cellcolor{red!25}0.995  
   & \cellcolor{red!25}0.990  
   & \cellcolor{red!25}0.991  
   & \cellcolor{orange!25}0.974  
   \\
 &  & LPIPS 
   & \cellcolor{red!25}0.040  
   & \cellcolor{red!25}0.031  
   & 0.088
   & \cellcolor{red!25}0.026  
   & \cellcolor{red!25}0.014  
   & \cellcolor{red!25}0.019  
   & \cellcolor{red!25}0.039  
   & \cellcolor{red!25}0.039  
   & 0.064
   \\
\cline{2-12}
\hline
\end{tabular}}
\label{tab:map_replica}
\end{table*}}

\subsubsection{Replica}
Among monocular systems, SplatMap achieved an average PSNR of 36.864, outperforming GLORIE-SLAM (31.04) by 18.8\% and PhotoSLAM (33.303) by 10.7\%. Similarly, our SSIM score of 0.985 represents a 1.5\% improvement over GLORIE-SLAM (0.97) and a 6.4\% improvement over PhotoSLAM (0.926). In perceptual quality (LPIPS), SplatMap achieved 0.040, which is a 66.7\% reduction compared to GLORIE-SLAM (0.12) and a 49.4\% reduction compared to PhotoSLAM (0.079).

When compared to RGB-D systems, SplatMap consistently demonstrated superior results despite using only monocular input. For example, it achieved a higher PSNR than NICE-SLAM (24.420) and E-SLAM (29.080), with relative improvements of 50.9\% and 26.8\%, respectively.

\begin{table*}[!t]
\centering
\caption{PSNR, SSIM, and LPIPS Metrics for Different SLAM Methods on TUM-RGBD dataset.}
\resizebox{.7\textwidth}{!}{
\begin{tabular}{|c|c|c|c|c|c|c|c|c|}
\cline{1-9}
 Input & Method & Metric & AVE & f1/desk & f1/desk2 & f1/room & f2/xyz & f3/off\\
\cline{1-9}
\multirow{3}{*}{RGBD} 
 & \multirow{3}{*}{SplaTAM} 
 & PSNR & - & \cellcolor{orange!25} 22.00 & - & - & \cellcolor{orange!25} 24.50 &\cellcolor{orange!25}  21.90 \\
 &  & SSIM & - & \cellcolor{orange!25} 0.86 & - & - & \cellcolor{red!25} 0.95 &\cellcolor{orange!25}  0.88 \\
 &  & LPIPS & - & \cellcolor{orange!25} 0.23 & - & - & \cellcolor{orange!25} 0.10 & \cellcolor{orange!25} 0.20 \\
\cline{1-9}
\multirow{12}{*}{MONO} & \multirow{3}{*}{MonoGS} 
 & PSNR & \cellcolor{yellow!25}18.81 & 19.67 & \cellcolor{orange!25} 19.16 &\cellcolor{yellow!25}18.41 & 16.17 & 20.63  \\
 &  & SSIM & \cellcolor{yellow!25}0.70 & 0.73 & \cellcolor{yellow!25}0.66 &\cellcolor{yellow!25} 0.64 & 0.72 & \cellcolor{yellow!25} 0.77  \\
 &  & LPIPS & \cellcolor{yellow!25}0.39 & 0.33 & \cellcolor{yellow!25}0.48 &\cellcolor{yellow!25} 0.51 & 0.31 & 0.34  \\
\cline{2-9}& \multirow{3}{*}{GIORIE-SLAM} 
 & PSNR & \cellcolor{orange!25}20.99 & 20.26 &\cellcolor{yellow!25} 19.09 &\cellcolor{orange!25}  18.78 & \cellcolor{red!25}25.62 & \cellcolor{yellow!25}21.21  \\
 &  & SSIM &\cellcolor{orange!25} 0.77 & \cellcolor{yellow!25}0.79 & \cellcolor{orange!25} 0.92 &\cellcolor{orange!25}  0.73 & 0.72 & 0.72 \\
 &  & LPIPS &\cellcolor{orange!25} 0.30 &\cellcolor{yellow!25} 0.31 & \cellcolor{orange!25} 0.38 &\cellcolor{orange!25}  0.38 & \cellcolor{red!25} 0.09 & 0.32  \\
\cline{2-9}& \multirow{3}{*}{PhotoSLAM} 
 & PSNR & - & \cellcolor{yellow!25}20.97 & - & - & 21.07 & 19.59  \\
 &  & SSIM & - & 0.74 & - & - & \cellcolor{yellow!25}0.73 & 0.69  \\
 &  & LPIPS & - & \cellcolor{orange!25} 0.23 & - & - & 0.17 &\cellcolor{yellow!25} 0.24  \\
\cline{2-9}
 & \multirow{3}{*}{Ours} 
 & PSNR & \cellcolor{red!25}23.121 & \cellcolor{red!25}22.414 & \cellcolor{red!25}24.194 & \cellcolor{red!25}22.501 &\cellcolor{yellow!25} 22.791 & \cellcolor{red!25}23.707 \\
 &  & SSIM &\cellcolor{red!25} 0.879 & \cellcolor{red!25}0.900 & \cellcolor{red!25}0.842 & \cellcolor{red!25}0.862 & \cellcolor{orange!25} 0.892 &\cellcolor{red!25} 0.900 \\
 &  & LPIPS & \cellcolor{red!25}0.196 & \cellcolor{red!25}0.188 &\cellcolor{red!25} 0.251 & \cellcolor{red!25}0.229 &\cellcolor{yellow!25} 0.156 &\cellcolor{red!25} 0.154  \\
\cline{1-9}
\end{tabular}
}
\label{tab:map_tum}
\end{table*}

\subsubsection{TUM}
\begin{figure*}[!ht]
    \centering
    \includegraphics[width=0.75\linewidth]{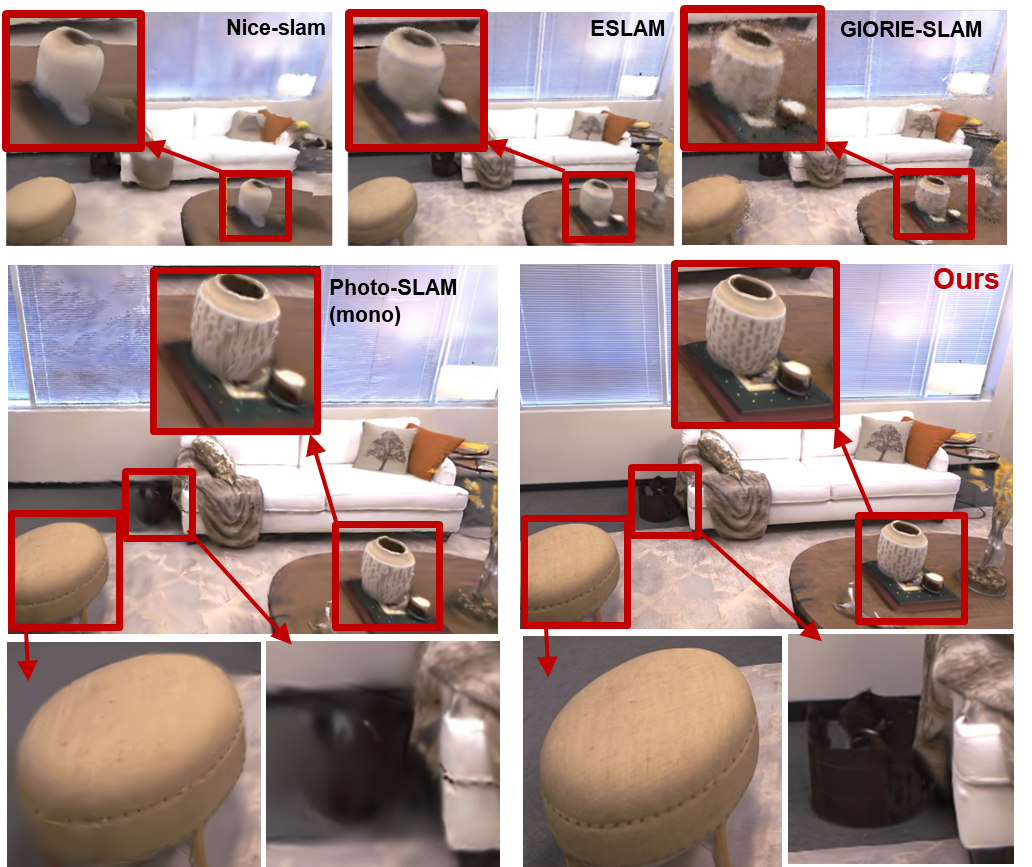}
    \caption{Mapping Comparison On Replica Dataset}
    \label{fig:map_vis_re}
\end{figure*}

\begin{figure}[!ht]
    \centering
    \includegraphics[width=\linewidth]{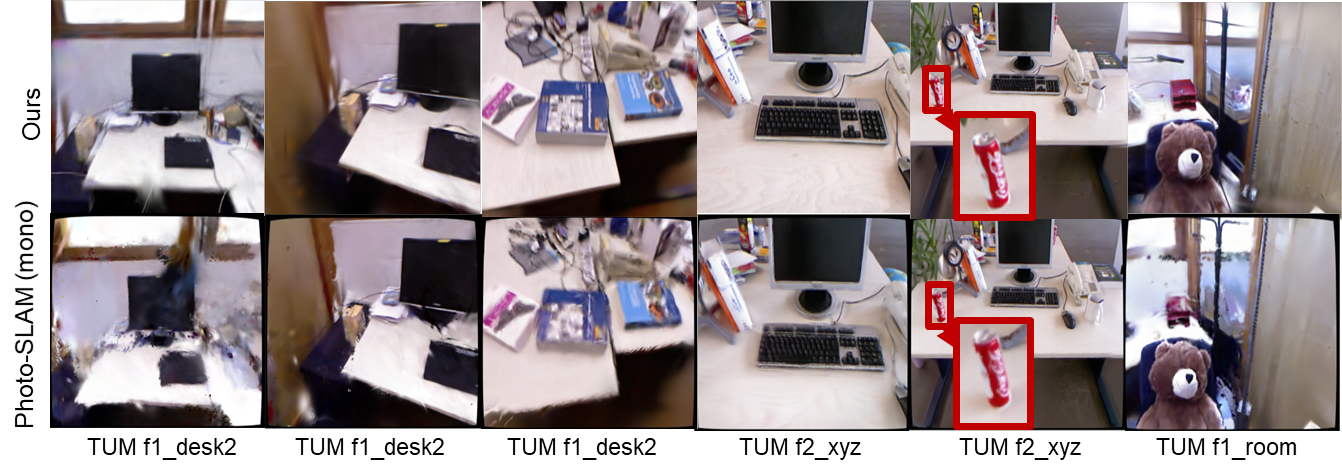}
    \caption{Mapping Comparison On TUM dataset}
    \label{fig:map_vis_tuml}
\end{figure}

On the TUM-RGBD dataset, SplatMap achieved an average PSNR of 23.121, surpassing GLORIE-SLAM (20.99) by 10.2\% and PhotoSLAM (19.53) by 18.3\%. The SSIM of 0.879 showed a 14.1\% improvement over PhotoSLAM (0.77) and a 6.6\% improvement over GLORIE-SLAM (0.82). In LPIPS, SplatMap achieved 0.196, a 34.7\% reduction compared to GLORIE-SLAM (0.30) and 14.8\% lower than PhotoSLAM (0.23). In Fig.~\ref{fig:map_vis_re} and~\ref{fig:map_vis_tuml}, we visualize the detailed comparisons between our method and SOTA SLAM systems on the Replica and TUM-RGBD datasets, respectively. 

\subsection{Runtime Analysis}

We assess the near real-time performance of our algorithm in Tab.~\ref{tab:resource}. We benchmark the runtime on an NVIDIA RTX A6000 GPU with 48 GB of memory. Our system operates at an average frame rate of 3.74 fps, which, while lower than that of GO-SLAM~\cite{zhang2023go}and ESLAM~\cite{johari2023eslamefficientdenseslam}, significantly outperforms Nice-SLAM~\cite{zhu2022nice}, GlORIE-SLAM~\cite{zhang2024glorie}, and MonoGS~\cite{matsuki2024gaussian}. The peak GPU memory consumption of our method is 14.96 GiB, which is competitive with existing state-of-the-art approaches. These results highlight that our method offers a balanced trade-off between computational efficiency and mapping accuracy, making it viable for online applications.

\begin{table}
    \centering
    \caption{ Memory and Running Time Evaluation on Replica room0}
    \resizebox{.8\textwidth}{!}{
    \begin{tabular}{|c|c|c|c|c|c|c|}
    \hline
         & ESLAM & Nice-SLAM & GO-SLAM & GlORIE-SLAM & MonoGS & Ours \\
         \hline
        Avg. FPS& 5.55 &0.48  & 8.36 & 0.23 & 0.32 & 3.74\\
        \hline
        GPU Usage (GiB) & - & - & 18.50 & 15.22 & 14.62  & 14.96\\
        \hline
    \end{tabular}}
    \label{tab:resource}
\end{table}

\subsection{Ablation Study}

\begin{table*}[ht]
\centering
\caption{Ablation Study on Replica and TUM-RGBD Datasets}
\resizebox{.8\textwidth}{!}{
\begin{tabular}{|c|c|c|c|c|c|c|c|c|c|c|}
\cline{1-11}
\multicolumn{5}{|c|}{Ablation Components} & \multicolumn{3}{|c|}{Replica-r0} & \multicolumn{3}{|c|}{TUM-RGBD-f1/desk2} \\
\cline{1-11}
&Base & SMOOTH & SIAD & SSIM & PSNR & SSIM & LPIPS & PSNR & SSIM & LPIPS \\
\cline{1-11}
\textcolor{red}A & \textcolor{green}{$\checkmark$} & \textcolor{red}{$\times$}  & \textcolor{red}{$\times$}  & \textcolor{red}{$\times$}  & 33.344 & 0.979 & 0.054 &  23.339   & 0.871    & 0.219    \\
\textcolor{red}B & \textcolor{green}{$\checkmark$} & \textcolor{red}{$\times$}  & \textcolor{green}{$\checkmark$} & \textcolor{red}{$\times$}  & 33.943 & 0.982 & 0.046 & 23.362   & 0.866   & 0.213    \\
\textcolor{red}C & \textcolor{green}{$\checkmark$} & \textcolor{green}{$\checkmark$}  & \textcolor{red}{$\times$}  & \textcolor{red}{$\times$}  & \cellcolor{yellow!25}34.444 & \cellcolor{yellow!25}0.985 & \cellcolor{yellow!25}0.037 & \cellcolor{yellow!25}23.951    &  \cellcolor{yellow!25}0.883    & \cellcolor{orange!25}0.190    \\
\textcolor{red}D & \textcolor{green}{$\checkmark$} & \textcolor{green}{$\checkmark$}  & \textcolor{green}{$\checkmark$} & \textcolor{red}{$\times$}  & \cellcolor{orange!25}34.549 & \cellcolor{orange!25}0.985 & \cellcolor{orange!25}0.036 & \cellcolor{orange!25}24.104    & \cellcolor{orange!25}0.888    & \cellcolor{yellow!25}0.191    \\
\textcolor{red}E & \textcolor{green}{$\checkmark$} & \textcolor{green}{$\checkmark$}  & \textcolor{green}{$\checkmark$} & \textcolor{green}{$\checkmark$} & \cellcolor{red!25}34.693 & \cellcolor{red!25}0.986 & \cellcolor{red!25}0.033 & \cellcolor{red!25}24.413    & \cellcolor{red!25} 0.896    & \cellcolor{red!25}0.179    \\
\cline{1-11}
\end{tabular}}
\label{tab:abl}
\end{table*}

\begin{figure}
    \centering
    \includegraphics[width=.8\linewidth]{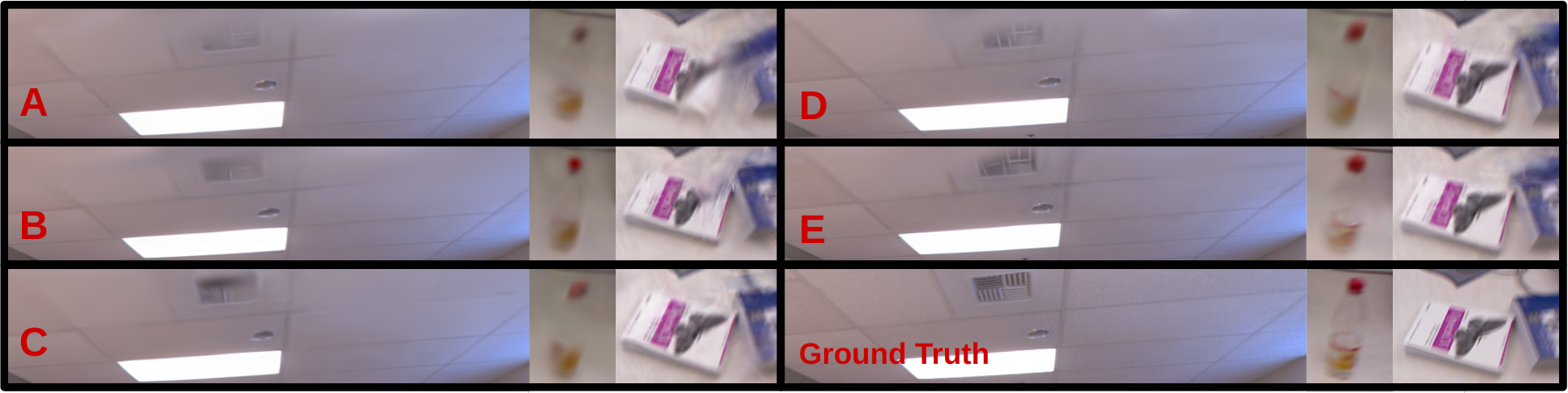}
    \caption{Ablation Study Visualization. The labels correspond to the respective experimental settings in Tab.~\ref{tab:abl}}
    \label{fig:abl}
\end{figure}

Table~\ref{tab:abl} presents the results of the ablation study on the Replica Dataset Room0 and TUM-RGBD dataset (f1/desk2), evaluating the impact of different components added to the base model. 
The Base model in our work refers to a vanilla online Gaussian SLAM system that integrates DROID-SLAM with 3DGS and incorporates an edge-aware loss to improve structural consistency. Here, “SMOOTH” refers to our proposed smooth weighting function (introduced in Sec.~\ref{sec:keyframe}, Equ.~\ref{equ:smooth}), which modulates the edge-aware loss using a Gaussian-like function to balance edge emphasis with spatial continuity. The configuration denoted as "Base + SIAD + SMOOTH + MS-SSIM" represents the full experimental setup used throughout our paper. In particular, this exact configuration is employed to produce the results shown in Fig.~\ref{fig:map_vis_re} and Fig.~\ref{fig:map_vis_tuml}

The Base model achieves a PSNR of 33.344, SSIM of 0.979, and LPIPS of 0.054. By incorporating SIAD, the PSNR improves to 33.943, and both SSIM (0.982) and LPIPS (0.046) metrics indicate enhanced image quality, demonstrating the effectiveness of SIAD in refining the reconstruction. Adding the SMOOTH component improves the PSNR to 34.444, SSIM to 0.985, and LPIPS to 0.037, indicating that smoothness constraints significantly enhance geometric consistency and rendering quality. Combining SMOOTH and SIAD further boosts the PSNR to 34.549, while maintaining a high SSIM of 0.985 and a slightly improved LPIPS of 0.036. Finally, integrating MS-SSIM (Multi-Scale Structural Similarity) yields the best results, with a PSNR of 34.693, SSIM of 0.986, and LPIPS of 0.033.

 The experiments on the TUM-RGBD dataset exhibit a similar trend: incorporating SIAD, SMOOTH, and MS-SSIM consistently improves the reconstruction quality. This consistency across two distinct datasets demonstrates that the improvements observed in our ablation study are generalizable and effective across diverse scenes.

The observed improvements in the ablation study are a direct consequence of our architectural innovations. Specifically, the SIAD module refines the dense point cloud by leveraging SLAM’s dynamic updates to prune erroneous points and densify only those that are newly reliable, thereby reducing ghosting artifacts and improving geometric accuracy. The SMOOTH component—our proposed smooth weighting function—modulates the edge-aware loss to prevent over-penalization in high-gradient regions while maintaining spatial continuity, which enhances the overall rendering consistency. Moreover, the incorporation of MS-SSIM into the composite RGB loss ensures multi-scale perceptual alignment, capturing structural details more robustly than single-scale SSIM. Together, these components synergistically contribute to significant gains in PSNR, SSIM, and LPIPS, as demonstrated consistently across both the Replica and TUM-RGBD datasets.

\section{Conclusion and Future Work}
In this paper, we introduced a novel framework for high-fidelity monocular 3D reconstruction by integrating dense SLAM with 3DGS. Our approach addresses the limitations of traditional sparse reconstruction pipelines by leveraging SLAM's dense depth and pose estimation capabilities to densify Gaussian scene representations dynamically. The proposed \textit{SLAM-Informed Adaptive Densification} transforms conventional sparse SLAM point cloud based Gaussian splatting into a dense reconstruction process, enabling richer geometric detail. Additionally, the \textit{Geometry-Guided Optimization} framework incorporates edge-aware geometric constraints, ensuring both structural accuracy and visual fidelity in the reconstructed scenes.

The effectiveness of our framework was demonstrated through extensive experiments on both the Replica and TUM-RGBD datasets, where our system achieved state-of-the-art performance among monocular SLAM systems. Specifically, on the Replica dataset, our method achieved a PSNR of 36.864, SSIM of 0.985, and LPIPS of 0.040, representing improvements of 10.7\%, 6.4\%, and 49.4\%, respectively, over Photo-SLAM, the previous SOTA monocular method. On the TUM-RGBD dataset, our method achieved a PSNR of 23.121, SSIM of 0.879, and LPIPS of 0.196, surpassing GLORIE-SLAM by 10.2\%, 6.6\%, and 34.7\%, respectively.

Notably, despite being a monocular system, our framework demonstrated competitive performance compared to RGB-D SLAM systems, such as NICE-SLAM and SplaTAM, while maintaining real-time inference capabilities. By bridging the gap between sparse and dense 3D scene representations, our approach sets a new benchmark for monocular dense 3D reconstruction pipelines. Future research directions include expanding the scalability of the system to handle larger and more dynamic environments, exploring the integration of semantic understanding for object-aware mapping and interaction, and investigating further optimizations to reduce memory and computational requirements for resource-constrained devices. These advancements will further enhance the practicality and applicability of our system for real-world applications.

While our current experiments focus on indoor scenarios, we recognize that outdoor environments present additional challenges, such as variable lighting conditions and larger-scale scenes. Moreover, the performance on challenging customer sequences still leaves room for improvement compared to conventional 3DGS. In future work, we plan to extend our framework to custom datasets collected under diverse conditions and to outdoor scenarios. This extension will involve investigating robustness across different devices and acquisition settings, ultimately aiming to enhance performance in these more challenging contexts.
\begin{acks}
The authors would like to thank our primary sponsors of this research: Mr. Clayton Burford
of the Battlespace Content Creation (BCC) team at Simulation and Training Technology
Center (STTC). This work is supported by University Affiliated Research Center (UARC)
award W911NF-14-D-0005. Statements and opinions expressed and content included do not
necessarily reflect the position or the policy of the Government, and no official endorsement
should be inferred.
\end{acks}

\newpage
{
    \small
    \bibliographystyle{ACM-Reference-Format}
    \bibliography{main}


\begin{thebibliography}{35}


\ifx \showCODEN    \undefined \def \showCODEN     #1{\unskip}     \fi
\ifx \showDOI      \undefined \def \showDOI       #1{#1}\fi
\ifx \showISBNx    \undefined \def \showISBNx     #1{\unskip}     \fi
\ifx \showISBNxiii \undefined \def \showISBNxiii  #1{\unskip}     \fi
\ifx \showISSN     \undefined \def \showISSN      #1{\unskip}     \fi
\ifx \showLCCN     \undefined \def \showLCCN      #1{\unskip}     \fi
\ifx \shownote     \undefined \def \shownote      #1{#1}          \fi
\ifx \showarticletitle \undefined \def \showarticletitle #1{#1}   \fi
\ifx \showURL      \undefined \def \showURL       {\relax}        \fi
\providecommand\bibfield[2]{#2}
\providecommand\bibinfo[2]{#2}
\providecommand\natexlab[1]{#1}
\providecommand\showeprint[2][]{arXiv:#2}

\bibitem[Engel et~al\mbox{.}(2014)]%
        {engel2014lsd}
\bibfield{author}{\bibinfo{person}{Jakob Engel}, \bibinfo{person}{Thomas Sch{\"o}ps}, {and} \bibinfo{person}{Daniel Cremers}.} \bibinfo{year}{2014}\natexlab{}.
\newblock \showarticletitle{LSD-SLAM: Large-scale direct monocular SLAM}. In \bibinfo{booktitle}{\emph{European conference on computer vision}}. Springer, \bibinfo{pages}{834--849}.
\newblock


\bibitem[Fisher et~al\mbox{.}(2021)]%
        {fisher2021colmap}
\bibfield{author}{\bibinfo{person}{Alex Fisher}, \bibinfo{person}{Ricardo Cannizzaro}, \bibinfo{person}{Madeleine Cochrane}, \bibinfo{person}{Chatura Nagahawatte}, {and} \bibinfo{person}{Jennifer~L Palmer}.} \bibinfo{year}{2021}\natexlab{}.
\newblock \showarticletitle{ColMap: A memory-efficient occupancy grid mapping framework}.
\newblock \bibinfo{journal}{\emph{Robotics and Autonomous Systems}}  \bibinfo{volume}{142} (\bibinfo{year}{2021}), \bibinfo{pages}{103755}.
\newblock


\bibitem[Huang et~al\mbox{.}(2024b)]%
        {huang20242d}
\bibfield{author}{\bibinfo{person}{Binbin Huang}, \bibinfo{person}{Zehao Yu}, \bibinfo{person}{Anpei Chen}, \bibinfo{person}{Andreas Geiger}, {and} \bibinfo{person}{Shenghua Gao}.} \bibinfo{year}{2024}\natexlab{b}.
\newblock \showarticletitle{2d gaussian splatting for geometrically accurate radiance fields}. In \bibinfo{booktitle}{\emph{ACM SIGGRAPH 2024 conference papers}}. \bibinfo{pages}{1--11}.
\newblock


\bibitem[Huang et~al\mbox{.}(2024a)]%
        {huang2024photo}
\bibfield{author}{\bibinfo{person}{Huajian Huang}, \bibinfo{person}{Longwei Li}, \bibinfo{person}{Hui Cheng}, {and} \bibinfo{person}{Sai-Kit Yeung}.} \bibinfo{year}{2024}\natexlab{a}.
\newblock \showarticletitle{Photo-SLAM: Real-time Simultaneous Localization and Photorealistic Mapping for Monocular Stereo and RGB-D Cameras}. In \bibinfo{booktitle}{\emph{Proceedings of the IEEE/CVF Conference on Computer Vision and Pattern Recognition}}. \bibinfo{pages}{21584--21593}.
\newblock


\bibitem[Johari et~al\mbox{.}(2023)]%
        {johari2023eslamefficientdenseslam}
\bibfield{author}{\bibinfo{person}{Mohammad~Mahdi Johari}, \bibinfo{person}{Camilla Carta}, {and} \bibinfo{person}{François Fleuret}.} \bibinfo{year}{2023}\natexlab{}.
\newblock \bibinfo{title}{ESLAM: Efficient Dense SLAM System Based on Hybrid Representation of Signed Distance Fields}.
\newblock
\newblock
\showeprint[arxiv]{2211.11704}~[cs.CV]
\urldef\tempurl%
\url{https://arxiv.org/abs/2211.11704}
\showURL{%
\tempurl}


\bibitem[Keetha et~al\mbox{.}(2024)]%
        {keetha2024splatam}
\bibfield{author}{\bibinfo{person}{Nikhil Keetha}, \bibinfo{person}{Jay Karhade}, \bibinfo{person}{Krishna~Murthy Jatavallabhula}, \bibinfo{person}{Gengshan Yang}, \bibinfo{person}{Sebastian Scherer}, \bibinfo{person}{Deva Ramanan}, {and} \bibinfo{person}{Jonathon Luiten}.} \bibinfo{year}{2024}\natexlab{}.
\newblock \showarticletitle{SplaTAM: Splat Track \& Map 3D Gaussians for Dense RGB-D SLAM}. In \bibinfo{booktitle}{\emph{Proceedings of the IEEE/CVF Conference on Computer Vision and Pattern Recognition}}. \bibinfo{pages}{21357--21366}.
\newblock


\bibitem[Kerbl et~al\mbox{.}(2023)]%
        {kerbl20233d}
\bibfield{author}{\bibinfo{person}{Bernhard Kerbl}, \bibinfo{person}{Georgios Kopanas}, \bibinfo{person}{Thomas Leimk{\"u}hler}, {and} \bibinfo{person}{George Drettakis}.} \bibinfo{year}{2023}\natexlab{}.
\newblock \showarticletitle{3D Gaussian Splatting for Real-Time Radiance Field Rendering.}
\newblock \bibinfo{journal}{\emph{ACM Trans. Graph.}} \bibinfo{volume}{42}, \bibinfo{number}{4} (\bibinfo{year}{2023}), \bibinfo{pages}{139--1}.
\newblock


\bibitem[Liu et~al\mbox{.}(2024)]%
        {liu2024atomgs}
\bibfield{author}{\bibinfo{person}{Rong Liu}, \bibinfo{person}{Rui Xu}, \bibinfo{person}{Yue Hu}, \bibinfo{person}{Meida Chen}, {and} \bibinfo{person}{Andrew Feng}.} \bibinfo{year}{2024}\natexlab{}.
\newblock \showarticletitle{AtomGS: Atomizing Gaussian Splatting for High-Fidelity Radiance Field}.
\newblock \bibinfo{journal}{\emph{arXiv preprint arXiv:2405.12369}} (\bibinfo{year}{2024}).
\newblock


\bibitem[Matsuki et~al\mbox{.}(2024)]%
        {matsuki2024gaussian}
\bibfield{author}{\bibinfo{person}{Hidenobu Matsuki}, \bibinfo{person}{Riku Murai}, \bibinfo{person}{Paul~HJ Kelly}, {and} \bibinfo{person}{Andrew~J Davison}.} \bibinfo{year}{2024}\natexlab{}.
\newblock \showarticletitle{Gaussian splatting slam}. In \bibinfo{booktitle}{\emph{Proceedings of the IEEE/CVF Conference on Computer Vision and Pattern Recognition}}. \bibinfo{pages}{18039--18048}.
\newblock


\bibitem[Mildenhall et~al\mbox{.}(2021)]%
        {mildenhall2021nerf}
\bibfield{author}{\bibinfo{person}{Ben Mildenhall}, \bibinfo{person}{Pratul~P Srinivasan}, \bibinfo{person}{Matthew Tancik}, \bibinfo{person}{Jonathan~T Barron}, \bibinfo{person}{Ravi Ramamoorthi}, {and} \bibinfo{person}{Ren Ng}.} \bibinfo{year}{2021}\natexlab{}.
\newblock \showarticletitle{Nerf: Representing scenes as neural radiance fields for view synthesis}.
\newblock \bibinfo{journal}{\emph{Commun. ACM}} \bibinfo{volume}{65}, \bibinfo{number}{1} (\bibinfo{year}{2021}), \bibinfo{pages}{99--106}.
\newblock


\bibitem[Mur-Artal et~al\mbox{.}(2015)]%
        {mur2015orb}
\bibfield{author}{\bibinfo{person}{Raul Mur-Artal}, \bibinfo{person}{Jose Maria~Martinez Montiel}, {and} \bibinfo{person}{Juan~D Tardos}.} \bibinfo{year}{2015}\natexlab{}.
\newblock \showarticletitle{ORB-SLAM: a versatile and accurate monocular SLAM system}.
\newblock \bibinfo{journal}{\emph{IEEE transactions on robotics}} \bibinfo{volume}{31}, \bibinfo{number}{5} (\bibinfo{year}{2015}), \bibinfo{pages}{1147--1163}.
\newblock


\bibitem[Mur-Artal and Tard{\'o}s(2017)]%
        {mur2017orb}
\bibfield{author}{\bibinfo{person}{Raul Mur-Artal} {and} \bibinfo{person}{Juan~D Tard{\'o}s}.} \bibinfo{year}{2017}\natexlab{}.
\newblock \showarticletitle{Orb-slam2: An open-source slam system for monocular, stereo, and rgb-d cameras}.
\newblock \bibinfo{journal}{\emph{IEEE transactions on robotics}} \bibinfo{volume}{33}, \bibinfo{number}{5} (\bibinfo{year}{2017}), \bibinfo{pages}{1255--1262}.
\newblock


\bibitem[Pan et~al\mbox{.}(2024)]%
        {pan2024global}
\bibfield{author}{\bibinfo{person}{Linfei Pan}, \bibinfo{person}{D{\'a}niel Bar{\'a}th}, \bibinfo{person}{Marc Pollefeys}, {and} \bibinfo{person}{Johannes~L Sch{\"o}nberger}.} \bibinfo{year}{2024}\natexlab{}.
\newblock \showarticletitle{Global Structure-from-Motion Revisited}.
\newblock \bibinfo{journal}{\emph{arXiv preprint arXiv:2407.20219}} (\bibinfo{year}{2024}).
\newblock


\bibitem[Peng et~al\mbox{.}(2024)]%
        {peng2024rtg}
\bibfield{author}{\bibinfo{person}{Zhexi Peng}, \bibinfo{person}{Tianjia Shao}, \bibinfo{person}{Yong Liu}, \bibinfo{person}{Jingke Zhou}, \bibinfo{person}{Yin Yang}, \bibinfo{person}{Jingdong Wang}, {and} \bibinfo{person}{Kun Zhou}.} \bibinfo{year}{2024}\natexlab{}.
\newblock \showarticletitle{Rtg-slam: Real-time 3d reconstruction at scale using gaussian splatting}. In \bibinfo{booktitle}{\emph{ACM SIGGRAPH 2024 Conference Papers}}. \bibinfo{pages}{1--11}.
\newblock


\bibitem[Sandstr{\"o}m et~al\mbox{.}(2023)]%
        {sandstrom2023point}
\bibfield{author}{\bibinfo{person}{Erik Sandstr{\"o}m}, \bibinfo{person}{Yue Li}, \bibinfo{person}{Luc Van~Gool}, {and} \bibinfo{person}{Martin~R Oswald}.} \bibinfo{year}{2023}\natexlab{}.
\newblock \showarticletitle{Point-slam: Dense neural point cloud-based slam}. In \bibinfo{booktitle}{\emph{Proceedings of the IEEE/CVF International Conference on Computer Vision}}. \bibinfo{pages}{18433--18444}.
\newblock


\bibitem[Schops et~al\mbox{.}(2019)]%
        {schops2019bad}
\bibfield{author}{\bibinfo{person}{Thomas Schops}, \bibinfo{person}{Torsten Sattler}, {and} \bibinfo{person}{Marc Pollefeys}.} \bibinfo{year}{2019}\natexlab{}.
\newblock \showarticletitle{Bad slam: Bundle adjusted direct rgb-d slam}. In \bibinfo{booktitle}{\emph{Proceedings of the IEEE/CVF Conference on Computer Vision and Pattern Recognition}}. \bibinfo{pages}{134--144}.
\newblock


\bibitem[Straub et~al\mbox{.}(2019)]%
        {straub2019replicadatasetdigitalreplica}
\bibfield{author}{\bibinfo{person}{Julian Straub}, \bibinfo{person}{Thomas Whelan}, \bibinfo{person}{Lingni Ma}, \bibinfo{person}{Yufan Chen}, \bibinfo{person}{Erik Wijmans}, \bibinfo{person}{Simon Green}, \bibinfo{person}{Jakob~J. Engel}, \bibinfo{person}{Raul Mur-Artal}, \bibinfo{person}{Carl Ren}, \bibinfo{person}{Shobhit Verma}, \bibinfo{person}{Anton Clarkson}, \bibinfo{person}{Mingfei Yan}, \bibinfo{person}{Brian Budge}, \bibinfo{person}{Yajie Yan}, \bibinfo{person}{Xiaqing Pan}, \bibinfo{person}{June Yon}, \bibinfo{person}{Yuyang Zou}, \bibinfo{person}{Kimberly Leon}, \bibinfo{person}{Nigel Carter}, \bibinfo{person}{Jesus Briales}, \bibinfo{person}{Tyler Gillingham}, \bibinfo{person}{Elias Mueggler}, \bibinfo{person}{Luis Pesqueira}, \bibinfo{person}{Manolis Savva}, \bibinfo{person}{Dhruv Batra}, \bibinfo{person}{Hauke~M. Strasdat}, \bibinfo{person}{Renzo~De Nardi}, \bibinfo{person}{Michael Goesele}, \bibinfo{person}{Steven Lovegrove}, {and} \bibinfo{person}{Richard Newcombe}.}
  \bibinfo{year}{2019}\natexlab{}.
\newblock \bibinfo{title}{The Replica Dataset: A Digital Replica of Indoor Spaces}.
\newblock
\newblock
\showeprint[arxiv]{1906.05797}~[cs.CV]
\urldef\tempurl%
\url{https://arxiv.org/abs/1906.05797}
\showURL{%
\tempurl}


\bibitem[Sucar et~al\mbox{.}(2021)]%
        {sucar2021imap}
\bibfield{author}{\bibinfo{person}{Edgar Sucar}, \bibinfo{person}{Shikun Liu}, \bibinfo{person}{Joseph Ortiz}, {and} \bibinfo{person}{Andrew~J Davison}.} \bibinfo{year}{2021}\natexlab{}.
\newblock \showarticletitle{imap: Implicit mapping and positioning in real-time}. In \bibinfo{booktitle}{\emph{Proceedings of the IEEE/CVF international conference on computer vision}}. \bibinfo{pages}{6229--6238}.
\newblock


\bibitem[Tang et~al\mbox{.}(2023)]%
        {tang2023dreamgaussian}
\bibfield{author}{\bibinfo{person}{Jiaxiang Tang}, \bibinfo{person}{Jiawei Ren}, \bibinfo{person}{Hang Zhou}, \bibinfo{person}{Ziwei Liu}, {and} \bibinfo{person}{Gang Zeng}.} \bibinfo{year}{2023}\natexlab{}.
\newblock \showarticletitle{Dreamgaussian: Generative gaussian splatting for efficient 3d content creation}.
\newblock \bibinfo{journal}{\emph{arXiv preprint arXiv:2309.16653}} (\bibinfo{year}{2023}).
\newblock


\bibitem[Teed and Deng(2020)]%
        {teed2020raft}
\bibfield{author}{\bibinfo{person}{Zachary Teed} {and} \bibinfo{person}{Jia Deng}.} \bibinfo{year}{2020}\natexlab{}.
\newblock \showarticletitle{Raft: Recurrent all-pairs field transforms for optical flow}. In \bibinfo{booktitle}{\emph{Computer Vision--ECCV 2020: 16th European Conference, Glasgow, UK, August 23--28, 2020, Proceedings, Part II 16}}. Springer, \bibinfo{pages}{402--419}.
\newblock


\bibitem[Teed and Deng(2021)]%
        {teed2021droid}
\bibfield{author}{\bibinfo{person}{Zachary Teed} {and} \bibinfo{person}{Jia Deng}.} \bibinfo{year}{2021}\natexlab{}.
\newblock \showarticletitle{Droid-slam: Deep visual slam for monocular, stereo, and rgb-d cameras}.
\newblock \bibinfo{journal}{\emph{Advances in neural information processing systems}}  \bibinfo{volume}{34} (\bibinfo{year}{2021}), \bibinfo{pages}{16558--16569}.
\newblock


\bibitem[Tukan et~al\mbox{.}(2023)]%
        {tukan2023orbslam3}
\bibfield{author}{\bibinfo{person}{Murad Tukan}, \bibinfo{person}{Fares Fares}, \bibinfo{person}{Yotam Grufinkle}, \bibinfo{person}{Ido Talmor}, \bibinfo{person}{Loay Mualem}, \bibinfo{person}{Vladimir Braverman}, {and} \bibinfo{person}{Dan Feldman}.} \bibinfo{year}{2023}\natexlab{}.
\newblock \showarticletitle{Orbslam3-enhanced autonomous toy drones: Pioneering indoor exploration}.
\newblock \bibinfo{journal}{\emph{arXiv preprint arXiv:2312.13385}} (\bibinfo{year}{2023}).
\newblock


\bibitem[Vespa et~al\mbox{.}(2018)]%
        {vespa2018efficient}
\bibfield{author}{\bibinfo{person}{Emanuele Vespa}, \bibinfo{person}{Nikolay Nikolov}, \bibinfo{person}{Marius Grimm}, \bibinfo{person}{Luigi Nardi}, \bibinfo{person}{Paul~HJ Kelly}, {and} \bibinfo{person}{Stefan Leutenegger}.} \bibinfo{year}{2018}\natexlab{}.
\newblock \showarticletitle{Efficient octree-based volumetric SLAM supporting signed-distance and occupancy mapping}.
\newblock \bibinfo{journal}{\emph{IEEE Robotics and Automation Letters}} \bibinfo{volume}{3}, \bibinfo{number}{2} (\bibinfo{year}{2018}), \bibinfo{pages}{1144--1151}.
\newblock


\bibitem[Wang et~al\mbox{.}(2023)]%
        {wang2023co}
\bibfield{author}{\bibinfo{person}{Hengyi Wang}, \bibinfo{person}{Jingwen Wang}, {and} \bibinfo{person}{Lourdes Agapito}.} \bibinfo{year}{2023}\natexlab{}.
\newblock \showarticletitle{Co-slam: Joint coordinate and sparse parametric encodings for neural real-time slam}. In \bibinfo{booktitle}{\emph{Proceedings of the IEEE/CVF Conference on Computer Vision and Pattern Recognition}}. \bibinfo{pages}{13293--13302}.
\newblock


\bibitem[Wang et~al\mbox{.}(2003)]%
        {wang2003multiscale}
\bibfield{author}{\bibinfo{person}{Zhou Wang}, \bibinfo{person}{Eero~P Simoncelli}, {and} \bibinfo{person}{Alan~C Bovik}.} \bibinfo{year}{2003}\natexlab{}.
\newblock \showarticletitle{Multiscale structural similarity for image quality assessment}. In \bibinfo{booktitle}{\emph{The Thrity-Seventh Asilomar Conference on Signals, Systems \& Computers, 2003}}, Vol.~\bibinfo{volume}{2}. Ieee, \bibinfo{pages}{1398--1402}.
\newblock


\bibitem[Whelan et~al\mbox{.}(2015)]%
        {whelan2015elasticfusion}
\bibfield{author}{\bibinfo{person}{Thomas Whelan}, \bibinfo{person}{Stefan Leutenegger}, \bibinfo{person}{Renato~F Salas-Moreno}, \bibinfo{person}{Ben Glocker}, {and} \bibinfo{person}{Andrew~J Davison}.} \bibinfo{year}{2015}\natexlab{}.
\newblock \showarticletitle{ElasticFusion: Dense SLAM without a pose graph.}. In \bibinfo{booktitle}{\emph{Robotics: science and systems}}, Vol.~\bibinfo{volume}{11}. Rome, Italy, \bibinfo{pages}{3}.
\newblock


\bibitem[Wu et~al\mbox{.}(2024)]%
        {wu20244d}
\bibfield{author}{\bibinfo{person}{Guanjun Wu}, \bibinfo{person}{Taoran Yi}, \bibinfo{person}{Jiemin Fang}, \bibinfo{person}{Lingxi Xie}, \bibinfo{person}{Xiaopeng Zhang}, \bibinfo{person}{Wei Wei}, \bibinfo{person}{Wenyu Liu}, \bibinfo{person}{Qi Tian}, {and} \bibinfo{person}{Xinggang Wang}.} \bibinfo{year}{2024}\natexlab{}.
\newblock \showarticletitle{4d gaussian splatting for real-time dynamic scene rendering}. In \bibinfo{booktitle}{\emph{Proceedings of the IEEE/CVF Conference on Computer Vision and Pattern Recognition}}. \bibinfo{pages}{20310--20320}.
\newblock


\bibitem[Yan et~al\mbox{.}(2024a)]%
        {yan2024gsslamdensevisualslam}
\bibfield{author}{\bibinfo{person}{Chi Yan}, \bibinfo{person}{Delin Qu}, \bibinfo{person}{Dan Xu}, \bibinfo{person}{Bin Zhao}, \bibinfo{person}{Zhigang Wang}, \bibinfo{person}{Dong Wang}, {and} \bibinfo{person}{Xuelong Li}.} \bibinfo{year}{2024}\natexlab{a}.
\newblock \bibinfo{title}{GS-SLAM: Dense Visual SLAM with 3D Gaussian Splatting}.
\newblock
\newblock
\showeprint[arxiv]{2311.11700}~[cs.CV]
\urldef\tempurl%
\url{https://arxiv.org/abs/2311.11700}
\showURL{%
\tempurl}


\bibitem[Yan et~al\mbox{.}(2024b)]%
        {yan2024gs}
\bibfield{author}{\bibinfo{person}{Chi Yan}, \bibinfo{person}{Delin Qu}, \bibinfo{person}{Dan Xu}, \bibinfo{person}{Bin Zhao}, \bibinfo{person}{Zhigang Wang}, \bibinfo{person}{Dong Wang}, {and} \bibinfo{person}{Xuelong Li}.} \bibinfo{year}{2024}\natexlab{b}.
\newblock \showarticletitle{Gs-slam: Dense visual slam with 3d gaussian splatting}. In \bibinfo{booktitle}{\emph{Proceedings of the IEEE/CVF Conference on Computer Vision and Pattern Recognition}}. \bibinfo{pages}{19595--19604}.
\newblock


\bibitem[Yang et~al\mbox{.}(2022)]%
        {yang2022vox}
\bibfield{author}{\bibinfo{person}{Xingrui Yang}, \bibinfo{person}{Hai Li}, \bibinfo{person}{Hongjia Zhai}, \bibinfo{person}{Yuhang Ming}, \bibinfo{person}{Yuqian Liu}, {and} \bibinfo{person}{Guofeng Zhang}.} \bibinfo{year}{2022}\natexlab{}.
\newblock \showarticletitle{Vox-fusion: Dense tracking and mapping with voxel-based neural implicit representation}. In \bibinfo{booktitle}{\emph{2022 IEEE International Symposium on Mixed and Augmented Reality (ISMAR)}}. IEEE, \bibinfo{pages}{499--507}.
\newblock


\bibitem[Yu et~al\mbox{.}(2021)]%
        {yu2021plenoxels}
\bibfield{author}{\bibinfo{person}{Alex Yu}, \bibinfo{person}{Sara Fridovich-Keil}, \bibinfo{person}{Matthew Tancik}, \bibinfo{person}{Qinhong Chen}, \bibinfo{person}{Benjamin Recht}, {and} \bibinfo{person}{Angjoo Kanazawa}.} \bibinfo{year}{2021}\natexlab{}.
\newblock \showarticletitle{Plenoxels: Radiance fields without neural networks}.
\newblock \bibinfo{journal}{\emph{arXiv preprint arXiv:2112.05131}} \bibinfo{volume}{2}, \bibinfo{number}{3} (\bibinfo{year}{2021}), \bibinfo{pages}{6}.
\newblock


\bibitem[Zhang et~al\mbox{.}(2024)]%
        {zhang2024glorie}
\bibfield{author}{\bibinfo{person}{Ganlin Zhang}, \bibinfo{person}{Erik Sandstr{\"o}m}, \bibinfo{person}{Youmin Zhang}, \bibinfo{person}{Manthan Patel}, \bibinfo{person}{Luc Van~Gool}, {and} \bibinfo{person}{Martin~R Oswald}.} \bibinfo{year}{2024}\natexlab{}.
\newblock \showarticletitle{Glorie-slam: Globally optimized rgb-only implicit encoding point cloud slam}.
\newblock \bibinfo{journal}{\emph{arXiv preprint arXiv:2403.19549}} (\bibinfo{year}{2024}).
\newblock


\bibitem[Zhang et~al\mbox{.}(2023)]%
        {zhang2023go}
\bibfield{author}{\bibinfo{person}{Youmin Zhang}, \bibinfo{person}{Fabio Tosi}, \bibinfo{person}{Stefano Mattoccia}, {and} \bibinfo{person}{Matteo Poggi}.} \bibinfo{year}{2023}\natexlab{}.
\newblock \showarticletitle{Go-slam: Global optimization for consistent 3d instant reconstruction}. In \bibinfo{booktitle}{\emph{Proceedings of the IEEE/CVF International Conference on Computer Vision}}. \bibinfo{pages}{3727--3737}.
\newblock


\bibitem[Zhu et~al\mbox{.}(2024)]%
        {zhu2024nicer}
\bibfield{author}{\bibinfo{person}{Zihan Zhu}, \bibinfo{person}{Songyou Peng}, \bibinfo{person}{Viktor Larsson}, \bibinfo{person}{Zhaopeng Cui}, \bibinfo{person}{Martin~R Oswald}, \bibinfo{person}{Andreas Geiger}, {and} \bibinfo{person}{Marc Pollefeys}.} \bibinfo{year}{2024}\natexlab{}.
\newblock \showarticletitle{Nicer-slam: Neural implicit scene encoding for rgb slam}. In \bibinfo{booktitle}{\emph{2024 International Conference on 3D Vision (3DV)}}. IEEE, \bibinfo{pages}{42--52}.
\newblock


\bibitem[Zhu et~al\mbox{.}(2022)]%
        {zhu2022nice}
\bibfield{author}{\bibinfo{person}{Zihan Zhu}, \bibinfo{person}{Songyou Peng}, \bibinfo{person}{Viktor Larsson}, \bibinfo{person}{Weiwei Xu}, \bibinfo{person}{Hujun Bao}, \bibinfo{person}{Zhaopeng Cui}, \bibinfo{person}{Martin~R Oswald}, {and} \bibinfo{person}{Marc Pollefeys}.} \bibinfo{year}{2022}\natexlab{}.
\newblock \showarticletitle{Nice-slam: Neural implicit scalable encoding for slam}. In \bibinfo{booktitle}{\emph{Proceedings of the IEEE/CVF conference on computer vision and pattern recognition}}. \bibinfo{pages}{12786--12796}.
\newblock


\end{thebibliography}
}


\end{document}